%% file: main.tex
\documentclass{article}

\usepackage{arxiv}

\usepackage[utf8]{inputenc}
\usepackage[T1]{fontenc}
\usepackage{url}
\usepackage{booktabs}
\usepackage{amsfonts}
\usepackage{amsmath}
\usepackage{microtype}
\usepackage{graphicx}
\usepackage{natbib}
\usepackage{doi}
\usepackage{cleveref}

\usepackage{hyperref}

\usepackage{xcolor}
\usepackage{multirow}
\usepackage{float}
\usepackage{placeins}

\graphicspath{{Fig/}}

\newcommand{\botrule}{\bottomrule}

\newenvironment{tablenotes}{%
  \par\vspace{4pt}\footnotesize
  \renewcommand{\item}{\par\noindent}
}{%
  \par
}

\newcommand{\beginsupplement}{%
  \setcounter{table}{0}\renewcommand{\thetable}{S\arabic{table}}%
  \setcounter{figure}{0}\renewcommand{\thefigure}{S\arabic{figure}}%
  \newcounter{suppsec}\renewcommand{\thesuppsec}{S\arabic{suppsec}}%
  \newcounter{suppsubsec}[suppsec]\renewcommand{\thesuppsubsec}{\thesuppsec.\arabic{suppsubsec}}%
}

\usepackage{authblk}

\renewcommand{\shorttitle}{SNPgen: Phenotype-Supervised Synthetic Genotypes}

\title{SNPgen: Phenotype-Supervised Genotype Representation and Synthetic Data Generation via Latent Diffusion}

\author[1,2]{Andrea Lampis}
\author[2]{Michela Carlotta Massi}
\author[3]{Nicola Pirastu}
\author[4]{Francesca Ieva}
\author[1]{Matteo Matteucci}
\author[2,5]{Emanuele Di Angelantonio}

\affil[1]{DEIB, Politecnico di Milano, Milan 20133, Italy}
\affil[2]{Health Data Science Centre, Human Technopole, Milan 20157, Italy}
\affil[3]{Genomics Research Centre, Human Technopole, Milan 20157, Italy}
\affil[4]{MOX - Department of Mathematics, Politecnico di Milano, Milan 20133, Italy}
\affil[5]{Department of Public Health and Primary Care, University of Cambridge, Cambridge CB2 1TN, UK}

\date{}

\hypersetup{
  pdftitle={SNPgen: Phenotype-Supervised Genotype Representation and Synthetic Data Generation via Latent Diffusion},
  pdfauthor={Andrea Lampis, Michela Carlotta Massi, Nicola Pirastu, Francesca Ieva, Matteo Matteucci, Emanuele Di Angelantonio},
  pdfkeywords={synthetic genotypes, latent diffusion models, variational autoencoder, polygenic risk scores, privacy-preserving genomics}
}

\begin{document}

\maketitle

\begin{abstract}
\textbf{Motivation:} Polygenic risk scores and other genomic analyses require large individual-level genotype datasets, yet strict data access restrictions impede sharing. Synthetic genotype generation offers a privacy-preserving alternative, but most existing methods operate unconditionally---producing samples without phenotype alignment---or rely on unsupervised compression, creating a gap between statistical fidelity and downstream task utility. \\
\textbf{Results:} We present SNPgen, a two-stage conditional latent diffusion framework for generating phenotype-supervised synthetic genotypes. SNPgen combines GWAS-guided variant selection (1{,}024--2{,}048 trait-associated SNPs) with a variational autoencoder for genotype compression and a latent diffusion model conditioned on binary disease labels via classifier-free guidance. Evaluated on 458{,}724 UK Biobank individuals across four complex diseases (coronary artery disease, breast cancer, type~1 and type~2 diabetes), models trained on synthetic data matched real-data predictive performance in a train-on-synthetic, test-on-real protocol, approaching genome-wide PRS methods that use 2--6$\times$ more variants. Privacy analysis confirmed zero identical matches, near-random membership inference (AUC $\approx$ 0.50), preserved linkage disequilibrium structure, and high allele frequency correlation ($r \geq 0.95$) with source data. A controlled simulation with known causal effects verified faithful recovery of the imposed genetic association structure. \\
\textbf{Availability and implementation:} Code available at \url{https://github.com/ht-diva/SNPgen}. \\
\textbf{Contact:} \href{mailto:andrea.lampis@polimi.it}{andrea.lampis@polimi.it} \\
\textbf{Supplementary information:} Supplementary data are available in the Appendix.
\end{abstract}

\keywords{synthetic genotypes \and latent diffusion models \and variational autoencoder \and polygenic risk scores \and privacy-preserving genomics}

\section{Introduction}

Genome-wide association studies (GWAS) have identified thousands of genetic variants associated with complex diseases, enabling the development of polygenic risk scores (PRS) that combine small per-variant effects into clinically relevant risk stratification~\citep{choiTutorialGuidePerforming2020,priveLDpred2BetterFaster2021}. However, these analyses require large individual-level genotype datasets that are expensive to collect and subject to strict access restrictions~\citep{shringarpurePrivacyRisksGenomic2015,erlichRoutesBreachingProtecting2014}. Synthetic genotype generation offers a path forward: models trained on access-controlled data can produce shareable datasets that preserve population-level genetic structure without exposing individual genomes.

However, genotype generation at biobank scale is a high-dimensional problem:
genome-wide arrays contain hundreds of thousands to millions of variants linked
by population structure and long-range correlations, forcing most existing
generators to focus on reduced settings or rely on strong compression strategies.
Moreover, most SNP generators are unconditional, sampling genotypes from the population distribution without phenotype alignment. While useful for benchmarking or privacy-focused release, such synthetic cohorts are not immediately usable for supervised tasks without an added phenotype mechanism. Some pipelines jointly simulate phenotypes (e.g., HAPNEST~\citep{wharrieHAPNESTEfficientLargescale2023}), but via explicit genetic-architecture simulation rather than learning a phenotype-conditional genotype distribution.

In this work, we focus on phenotype-supervised genotype generation to produce synthetic samples that are immediately usable for downstream analyses. To address the challenges of genome-wide modelling, we introduce a GWAS-guided selection strategy that restricts modelling to trait-associated SNPs. Indeed, GWAS effect sizes provide a natural ranking for prioritising variants, and clumping-and-thresholding on this ranking is a standard strategy for constructing polygenic scores~\citep{choiPRSice2PolygenicRisk2019}. This concentrates model capacity on variants that carry trait-relevant signal and substantially reduces computational cost, offering a favourable trade-off between panel size and predictive power.

Building on this GWAS-guided representation, we present \textbf{SNPgen}, a two-stage conditional latent diffusion framework. In Stage~1, a Variational AutoEncoder (VAE) compresses one-hot encoded genotype sequences into a compact latent space. In Stage~2, a Latent Diffusion Model (LDM) conditioned on binary disease labels generates novel synthetic genotypes via iterative denoising with classifier-free guidance. Phenotype information thus enters at two levels: variant selection and the generative process itself.

Because SNPgen is designed for supervised downstream use, we evaluate it with utility and privacy-focused protocols. We first test SNPgen in a controlled simulation setting by simulating phenotypes using known causal effect sizes and verifying recovery of the expected associations, by performing GWAS on the generated synthetic data. Then, we validate our proposed approach on four complex traits with different polygenic architectures from the UK Biobank~\citep{sudlowUKBiobankOpen2015}--- coronary artery disease, breast cancer, type 1 diabetes, and type 2 diabetes---using a train-on-synthetic, test-on-real (TSTR) protocol, showing that our method preserves predictive performance comparable to real data, approaching that of genome-wide PRS methods that use 2--6$\times$ more variants. Finally, we demonstrate that the generated data provides strong privacy guarantees: zero identical matches, near-random membership inference, and high allele frequency correlation with the source population.
Together, these results position SNPgen as a practical, phenotype-supervised route to sharing task-ready synthetic genotype data at biobank scale.

\section{Related work}\label{sec:related}

Most existing genotype generators operate on restricted subsets of variants or on limited genomic segments~\citep{yelmenCreatingArtificialHuman2021,yelmenDeepConvolutionalConditional2023,ahronovizGenomeACGANEnhancingSynthetic2024}. The closest work to full-length generation is \citet{kennewegGeneratingSyntheticGenotypes2024}, who propose a diffusion framework for ``complete synthetic human genotypes''. However, scalability is achieved through a multi-stage representation pipeline inherited from \citet{luoPredictingPrevalenceComplex2023}: SNPs with GWAS $p > 0.05$ are first discarded (retaining ${\sim}$505\,K of ${\sim}$4.4\,M variants, much smaller than the whole genome) then per-gene PCAs compress the remaining variants into ${\sim}$75\,K concatenated components. This implies that (i) the subsequent diffusion model must operate on tens of thousands of PCA components, requiring large UNets and increasing both training cost and sampling latency; and (ii) PCA preserves directions of maximal genotype variance, often dominated by ancestry and population structure, rather than genotype--phenotype relationships, forcing the model to recover trait-relevant signal from a high-dimensional unsupervised representation. Indeed, the evaluation in \citet{kennewegGeneratingSyntheticGenotypes2024} is limited to rare-disease cohorts and ancestry classification, where genetic signals are comparatively strong, leaving open whether subtle polygenic effects underlying common complex traits are preserved.

Beyond scalability, \emph{phenotype readiness} is largely absent from the current landscape. The majority of generators are unconditional: they model the population genotype distribution and output samples without trait alignment~\citep{yelmenCreatingArtificialHuman2021,yelmenDeepConvolutionalConditional2023,wharrieHAPNESTEfficientLargescale2023}. Among the few conditioned approaches surveyed by \citet{kennewegGeneratingSyntheticGenotypes2024}, \citet{pereraGenerativeMomentMatching2022} condition on ancestry and population structure, while \citet{ahronovizGenomeACGANEnhancingSynthetic2024} generate genotypes tailored to subpopulations for population classification. Neither conditions on a clinical phenotype, so the resulting synthetic data are not directly usable for disease risk prediction or case/control modelling without an additional phenotype mechanism.

These gaps motivate SNPgen: rather than unconditional sampling or population-level conditioning, we combine GWAS-guided variant selection---as a lever for both scalability and signal concentration---with trait-conditioned generation, producing synthetic genotypes that are directly usable in supervised downstream tasks.

\section{Methods: SNPgen framework}\label{sec:methods}

SNPgen is a phenotype-supervised framework for generating task-ready synthetic genotypes. It combines (i) phenotype-guided variant selection to define a compact, trait-relevant genotype representation, and (ii) conditional latent diffusion to generate novel genotype samples aligned to a target phenotype. Given genotype data $\mathbf{G}$ and a phenotype label $y$ (binary in this work), SNPgen learns a conditional generator for a reduced SNP panel $\mathbf{G}_S$ selected using external GWAS evidence. The framework outputs synthetic genotypes $\tilde{\mathbf{G}}_S$ conditioned on $y$, designed to support supervised downstream modelling (e.g., risk prediction) while reducing computational cost relative to genome-wide generation.

\subsection{Phenotype-guided variant selection}\label{sec:snp_selection}

A key design choice in genotype generation is which variants to model. SNPgen adopts a phenotype-guided selection strategy based on external GWAS summary statistics for the trait of interest. Specifically, SNPs are ranked by GWAS significance (e.g., p-value) and pruned for redundancy using Linkage Disequilibrium (LD) clumping. From the clumped set, the top $L$ variants are retained to define the trait-specific SNP panel $S$ (with $L$ treated as a hyperparameter). This selection step concentrates modelling capacity on variants most informative for the target phenotype and reduces the dimensionality of the generative problem.

\subsection{Generative model}\label{sec:generative_model}

\begin{figure}[!t]
\centering
\includegraphics[width=\textwidth]{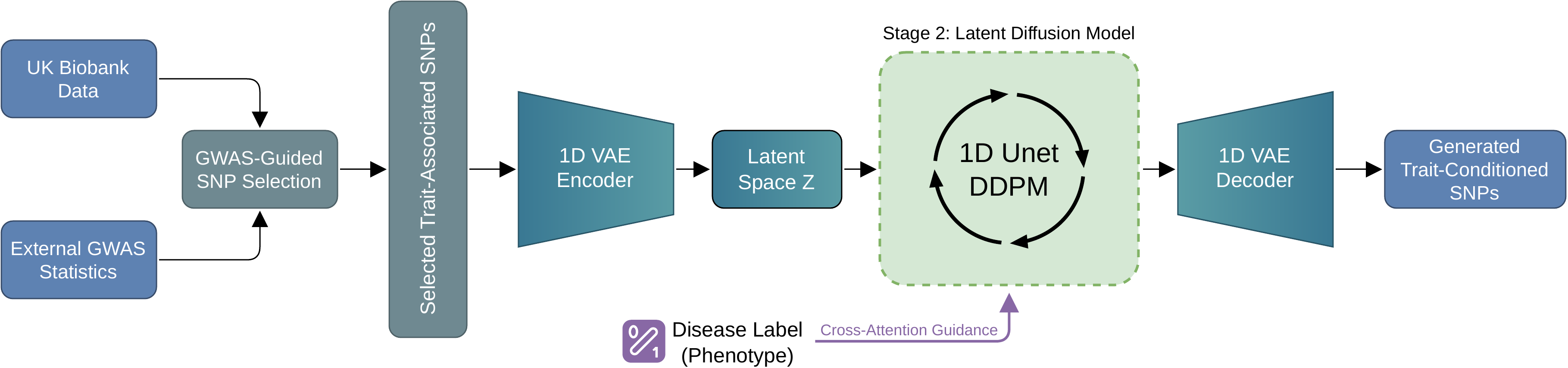}
\caption{SNPgen pipeline. UK Biobank genotypes and external GWAS summary statistics are combined for GWAS-guided SNP selection, producing a compact set of trait-associated variants. In Stage~1, a 1D VAE encoder compresses the selected SNPs into a continuous latent space~$\mathbf{z}$. In Stage~2, a latent diffusion model (1D UNet DDPM) generates novel latent representations conditioned on binary disease labels via cross-attention, which are then decoded into synthetic trait-conditioned genotypes.}\label{fig:architecture}
\end{figure}

SNPgen includes a two-stage generative model (Figure~\ref{fig:architecture}). Stage~1 uses a variational autoencoder (VAE)~\citep{kingmaAutoEncodingVariationalBayes2013} to compress discrete SNP sequences into a continuous latent space. Stage~2 trains a latent diffusion model (LDM)~\citep{rombachHighResolutionImageSynthesis2022} on the learned representations to generate novel synthetic genotypes.

\medskip
\noindent In Stage~1, the VAE architecture is adapted from the Stable Diffusion image autoencoder~\citep{rombachHighResolutionImageSynthesis2022}, replacing all 2D convolutions with 1D convolutions to process sequential genotype data. Each individual's genotype is encoded as a one-hot matrix $\mathbf{X} \in \{0,1\}^{3 \times L}$, where $L$ is the number of SNPs and the three channels represent homozygous reference, heterozygous, and homozygous alternative genotypes. The encoder is a 1D convolutional ResNet with five resolution levels and four downsampling stages. Latent vectors $\mathbf{z}$ are sampled via the standard Gaussian reparameterization trick. The decoder mirrors the encoder architecture, producing 3-channel logits over the genotype alphabet $\{0,1,2\}$.

The VAE is trained with a composite loss:
\begin{equation}\label{eq:vae_loss}
\mathcal{L}_\text{VAE} = \frac{\mathcal{L}_\text{recon}}{\exp(s)} + s + \beta \cdot D_\text{KL} + \lambda_\text{disc} \cdot \mathcal{L}_\text{adv},
\end{equation}
where $\mathcal{L}_\text{recon}$ is the per-position cross-entropy loss over genotype classes, $s$ is a learnable log-variance scalar (initialized to~0) that adaptively balances reconstruction and regularization, $D_\text{KL} = D_\text{KL}(q(\mathbf{z}|\mathbf{X}) \| p(\mathbf{z}))$ is the KL divergence with $\beta{=}1.0$, and $\mathcal{L}_\text{adv}$ is the loss from an auxiliary discriminator. The discriminator uses spectral normalization and hinge loss, and is activated after initial VAE stabilization; its weight $\lambda_\text{disc}$ is set adaptively via gradient-norm balancing.

\medskip
\noindent In Stage~2, the LDM operates on the VAE latent vectors $\mathbf{z} \in \mathbb{R}^{1 \times L/16}$ from the frozen, trained Stage~1 encoder. The denoiser is a 1D UNet, also adapted from Stable Diffusion, with three resolution levels and spatial transformer attention blocks at all scales (full specifications in Supplementary Section~\ref{sec:supp_architecture}). We employ the scaled-linear $\beta$ schedule from the LDM framework with 1{,}000 discretized timesteps and $\epsilon$-prediction~\citep{hoDenoisingDiffusionProbabilistic2020}. The training loss is the standard mean squared error between predicted and actual noise:
\begin{equation}\label{eq:diffusion_loss}
\mathcal{L}_\text{LDM} = \mathbb{E}_{t, \mathbf{z}_0, \boldsymbol{\epsilon}} \left[ \| \boldsymbol{\epsilon} - \boldsymbol{\epsilon}_\theta(\mathbf{z}_t, t, \mathbf{c}) \|_2^2 \right],
\end{equation}
where $\mathbf{z}_t$ is the noised latent, $t$ is the timestep, $\boldsymbol{\epsilon}_\theta$ is the denoiser, and $\mathbf{c}$ is the conditioning signal.

\medskip
\noindent Binary phenotype labels are embedded via a learned class embedder and injected through cross-attention in the UNet transformer blocks. Classifier-free guidance~\citep{hoClassifierFreeDiffusionGuidance2022} is applied during training.

\medskip
\noindent To generate synthetic genotypes conditioned on a target label $y$, SNPgen applies the trained diffusion model in reverse from Gaussian noise, yielding a synthetic latent $\tilde{\mathbf{z}}_0 \sim p(\mathbf{z}_0 \mid y)$. The frozen VAE decoder then maps $\tilde{\mathbf{z}}_0$ to a discrete genotype \mbox{$\tilde{\mathbf{G}}_S = \arg\max_{\mathbf{X}}\; p_\psi(\mathbf{X} \mid \tilde{\mathbf{z}}_0)$}, producing a synthetic SNP panel aligned to $y$. Sampling hyperparameters (denoising steps, sampler, guidance scale) are in Supplementary Section~\ref{sec:supp_training}.

\section{Evaluation}\label{sec:evaluation}

We evaluate SNPgen in the experimental setting along four complementary axes that match its goal of releasing task-ready synthetic genotypes: (i) genotype--phenotype signal preservation, (ii) downstream predictive utility, (iii) genomic fidelity (LD structure), and (iv) privacy risk.
First, we use a controlled simulation with known causal effects to test whether synthetic data preserve the imposed associations (via GWAS/effect recovery). Second, we measure utility on four real UK Biobank traits with a train-on-synthetic, test-on-real (TSTR) protocol~\citep{estebanRealvaluedMedicalTime2017}, and we benchmark against genome-wide PRS baselines to quantify how much predictive information is retained, or lost, by the GWAS-guided SNP selection. In this setting we also evaluate the potential of augmenting the number of cases via synthetic data generation. Third, we assess LD preservation to verify that local correlation structure is maintained beyond marginal allele frequencies. Finally, we quantify memorization and membership-inference leakage to ensure utility is not achieved by reproducing training individuals.

\subsection{Cohort and data}\label{sec:cohort}

We used genotype data from the UK Biobank~\citep{sudlowUKBiobankOpen2015}, a large-scale prospective cohort of approximately 500{,}000 individuals with genome-wide SNP data. Participants were selected based on self-reported white ancestry (Supplementary Section~\ref{sec:supp_phenotype}), yielding 458{,}724 individuals after ancestry-based filtering and intersection with available genotype data (${\sim}$97\,M imputed variants).

We considered four binary disease traits: coronary artery disease (CAD), breast cancer (BC; female participants only), type~1 diabetes (T1D), and type~2 diabetes (T2D). Phenotype definitions were derived from hospital episode statistics and self-reported diagnoses (full ICD-10/ICD-9 codes and self-report field references in Supplementary Section~\ref{sec:supp_phenotype}).

Data were split into training (70\%), validation (20\%), and test (10\%) sets via stratified random sampling to preserve phenotype proportions. Training set sizes ranged from 174{,}290 (BC, female-only) to 321{,}106 (CAD, T1D, T2D). Related individuals (up to third-degree relatedness, identified via KING kinship estimates provided by UK Biobank) were assigned to the same set to prevent information leakage through shared genetic background. All downstream performance metrics reported in this study are evaluated on the independent held-out test set.

For each trait, we applied LD clumping on external GWAS summary statistics (Supplementary Section~\ref{sec:supp_gwas}) using PLINK2~\citep{changSecondgenerationPLINKRising2015}, retaining the top $L$ SNPs by $p$-value: $L{=}2{,}048$ for CAD, T1D, and T2D and $L{=}1{,}024$ for BC; full details are in Supplementary Table~\ref{tab:snp_selection}.

For the controlled validation with known ground-truth genetic architecture, we additionally simulated a binary phenotype on the 2{,}048 CAD-selected SNPs. Per-SNP effect sizes were drawn from a three-component Gaussian mixture ($\pi = (0.3, 0.5, 0.2)$; $\mu = (0.8, 2.5, 0.0)$; $\sigma^2 = (0.2, 0.4, 0.1)$), with 5\% of SNPs set to zero effect (non-causal). Disease status was generated via a logistic model $p_i = \sigma(\beta_0 + \mathbf{G}_i \boldsymbol{\beta} + \varepsilon_i)$, where the intercept $\beta_0$ was optimised to target 30\% prevalence (achieved 32.9\%, 150{,}901 cases).

\subsection{Training}\label{sec:training}
Both stages were trained with the Adam optimizer and mixed-precision (FP16) training on a single NVIDIA Tesla V100 (32\,GB). The VAE ran for 400 epochs with exponential moving average (checkpoint: best validation reconstruction accuracy) and the LDM for 500 epochs with cosine learning rate schedule (checkpoint: lowest validation loss), each requiring approximately 24 hours. Full hyperparameters are in Supplementary Section~\ref{sec:supp_training}.

\subsection{Simulation Experiment}\label{sec:simulation_exp}

To directly test whether SNPgen preserves genotype--phenotype associations, we leverage the simulated trait where the ground-truth causal effects are known. After training SNPgen on the simulated labels, we generate a synthetic cohort matched in size and class balance to the real training set and run a standard GWAS (univariate logistic regression per SNP) on the synthetic data. We then compare the estimated per-SNP effects from synthetic data against the corresponding estimates on the real data, using correlation. Because GWAS is the same marginal association mechanism used to rank variants and is evaluated against known effects, this experiment provides a controlled, label-grounded validation that downstream utility reflects faithful recovery of the intended association structure rather than generative artefacts.

\subsection{Downstream risk prediction}\label{sec:downstream}

We assessed whether synthetic genotypes preserve disease-predictive signal using the four UK Biobank disease traits (CAD, BC, T1D, and T2D), with a TSTR protocol. In particular, for each trait, we compared four training conditions: Real (using real data in the original training set), Reconstructed (VAE encode--decode of the real training genotypes), Synthetic Matched (synthetic cohort matching the real training set size and class ratio), and Synthetic Augmented (class-balanced synthetic cohort of size $2 \times n_{\text{controls}}$, oversampling cases; sensitivity to augmentation proportion in Supplementary Section~\ref{sec:supp_augmentation}).
All predictive models trained on these four datasets were evaluated on the same held-out real test set (10\% of original data).
Because every model is tested on real genotypes, phenotype conditioning cannot artificially inflate downstream performance.

We evaluated multiple predictors to ensure results are not an artefact of a single downstream model. All models used genotype features only (no covariates) to isolate genotype-level predictive signal and ensure comparability across data conditions. Specifically, we compared three models using 5-fold stratified cross-validation:
\begin{itemize}
\item \textbf{XGBoost}: a strong nonlinear classifier able to exploit interactions among selected SNPs. The model was trained with grid search over 100 hyperparameter combinations and 5-fold inner CV (Supplementary Section~\ref{sec:supp_downstream_models}).
\item \textbf{XGBoost (Balanced):} XGBoost trained on a class-balanced subset (majority class downsampled to match the minority class).
\item \textbf{PRS Univariate:} a polygenic risk score where per-SNP effect sizes are estimated by performing a GWAS (i.e.\ univariate logistic regression) directly on the selected SNPs using the training cohort (rather than external summary statistics). The score is $\text{PRS} = \sum_i \hat{\beta}_i \cdot g_i$, where $g_i \in \{0, 1, 2\}$ is the minor allele dosage at SNP~$i$. The univariate PRS serves as a transparent linear baseline that directly reflects preservation of additive per-SNP effects.
\end{itemize}
An additional PRS computed from external GWAS summary statistics (\textbf{External PRS}) served as a reference for the performance of the above methods.

\smallskip
\noindent Finally, to contextualize predictive performance relative to standard whole-genome approaches and quantify the trade-off induced by GWAS-guided SNP selection, we evaluated two established PRS methods on the same UK Biobank cohort:
\begin{itemize}
\item \textbf{PRSice-2}~\citep{choiPRSice2PolygenicRisk2019}: clumping-and-thresholding PRS with 10 configurations (window $\in \{10, 50, 100, 250, 500\}$\,kb $\times$ $r^2 \in \{0.1, 0.5\}$). The best configuration per trait is reported; full results are in Supplementary Section~\ref{sec:supp_prsice}.
\item \textbf{LDpred2-auto}~\citep{priveLDpred2BetterFaster2021}: Bayesian genome-wide PRS using HapMap3+ LD reference (${\sim}$1.2\,M variants), 30 chains with 500 burn-in and 500 sampling iterations.
\end{itemize}
Both were evaluated using PRS-only logistic regression on the same genotype-only basis for comparability with SNPgen's approach.

\subsection{Privacy evaluation}\label{sec:privacy}

We evaluated privacy using six complementary metrics~\citep{yaleGenerationEvaluationPrivacy2020,sdmetrics,kennewegGeneratingSyntheticGenotypes2024}, applied to both reconstructed and synthetic data with the holdout test set as baseline. All distance computations used Hamming distance on one-hot encoded genotypes:

\begin{itemize}
\item \textbf{Identical Match Rate (IMR):} fraction of synthetic samples that are exact copies of training samples.
\item \textbf{Nearest Neighbour Distance Ratio (NNDR):} ratio $d_1 / d_2$ of distances to the first and second nearest training neighbours; values near~1.0 indicate no memorization.
\item \textbf{Membership Inference (MI):} AUC of a classifier distinguishing training from holdout samples based on distances to synthetic data; AUC~$\approx 0.5$ indicates no information leakage.
\item \textbf{Distance to Closest Record (DCR):} fraction of synthetic samples whose nearest training neighbour is closer than the 5th percentile of holdout--training distances.
\item \textbf{Nearest Neighbour Adversarial Accuracy (NNAA):} measures whether synthetic samples are systematically closer to training than holdout samples; computed on 50{,}000 randomly sampled training and synthetic individuals.
\item \textbf{MAF correlation ($r$):} Pearson correlation of minor allele frequencies between real and generated data; quantifies population-level allele frequency preservation.
\end{itemize}

\subsection{LD structure preservation}\label{sec:ld}

To assess whether SNPgen preserves linkage disequilibrium (LD) beyond marginal allele frequencies, we compared both the pairwise LD structure and the expected decay of LD with physical distance. We computed pairwise LD as $r^2$ (squared Pearson correlation of SNP dosages) in the original, VAE-reconstructed, and LDM-generated synthetic cohorts of all four UK Biobank traits, and compared the resulting LD matrices. We then quantified LD decay---in the original, reconstructed, and synthetic cohorts---by computing $r^2$ for each SNP pair together with their base-pair distance, and grouping pairs into 50 log-spaced distance bins. For each bin, we reported the mean $r^2$ ($\pm$ SEM) as a function of distance (details in Supplementary Section~\ref{sec:supp_ld}). Similar decay curves between real and generated data indicate that reconstructed and synthetic genotypes preserve the expected short-range LD and its decrease with genomic distance.

\section{Results and discussion}\label{sec:results}

\subsection{Simulation study}\label{sec:results_simulation}

PRS-univariate effect estimates from synthetic data show strong agreement with real-data betas (Pearson $r = 0.835$), substantially exceeding the agreement observed for VAE-reconstructed data ($r = 0.726$; Supplementary Figure~\ref{fig:supp_beta_sim}). This indicates that the phenotype-conditioned LDM preserves the marginal association structure more faithfully than unconditional VAE reconstruction in a setting with known ground-truth effects.

\subsection{Downstream risk prediction}\label{sec:results_downstream}

\begin{figure}[!t]
\centering
\includegraphics[width=\textwidth]{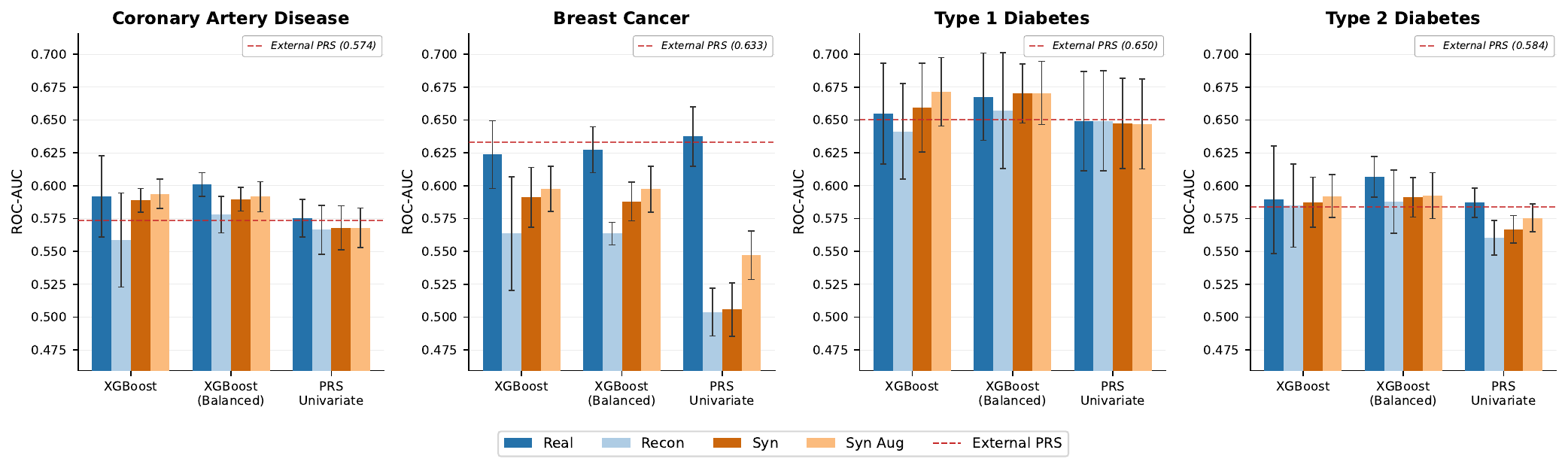}
\caption{Downstream risk prediction (ROC-AUC) across four UK Biobank traits. Bars show three models (XGBoost, XGBoost Balanced, PRS Univariate) under four data conditions: Real, Reconstructed, Synthetic, and Synthetic Augmented. Error bars: 95\% CI from 5-fold CV. Dashed lines: External PRS baseline. Cohort sizes: CAD $n{=}458{,}724$ (35{,}306 cases); BC $n{=}248{,}987$ (19{,}634 cases, female only); T1D $n{=}458{,}724$ (4{,}593 cases); T2D $n{=}458{,}724$ (38{,}541 cases).}\label{fig:downstream}
\end{figure}

Figure~\ref{fig:downstream} and Supplementary Table~\ref{tab:performance} summarize TSTR performance across all four traits, spanning concentrated (T1D, HLA-driven) to diffuse (CAD, T2D, BC) genetic architectures. This provides a stress test for whether selected, phenotype-conditioned synthetic genotypes retain trait-relevant predictive information, irrespective of the genetic architecture.

\smallskip
\noindent For T1D, synthetic data closely matched or exceeded real-data performance. XGBoost Balanced achieved $0.670 \pm 0.022$ on synthetic data versus $0.668 \pm 0.033$ on real data. The Synthetic Augmented condition yielded the highest AUC across all conditions ($0.671 \pm 0.026$ for XGBoost, $0.671 \pm 0.024$ for XGBoost Balanced). PRS Univariate remained stable across all conditions (0.647--0.649), consistent with the strong, concentrated genetic signal in the HLA region driving T1D risk.

\smallskip
\noindent In T2D prediction, Synthetic XGBoost achieved $0.587 \pm 0.019$, closely matching real data ($0.589 \pm 0.041$). XGBoost Balanced showed slightly larger gaps (real: $0.607 \pm 0.016$, synthetic: $0.591 \pm 0.015$). PRS performance degraded modestly from real ($0.587 \pm 0.011$) to synthetic ($0.567 \pm 0.011$), with Synthetic Augmented partially recovering performance ($0.576 \pm 0.011$).

\smallskip
\noindent Concerning CAD prediction, Synthetic Augmented achieved the highest non-real AUC (XGBoost: $0.594 \pm 0.011$; XGBoost Balanced: $0.592 \pm 0.011$), closely matching real-data performance (XGBoost: $0.592 \pm 0.031$; XGBoost Balanced: $0.601 \pm 0.009$). The matched Synthetic condition also performed well (XGBoost: $0.589 \pm 0.009$). PRS Univariate showed comparable performance across synthetic conditions (Synthetic: $0.568 \pm 0.017$; Syn Augmented: $0.568 \pm 0.015$) and reconstructed data ($0.566 \pm 0.019$).

\smallskip
\noindent Finally, in the BC cohort, Synthetic XGBoost retained approximately 95\% of real performance ($0.591 \pm 0.023$ vs.\ $0.624 \pm 0.026$). Synthetic Augmented slightly improved over matched synthetic ($0.598 \pm 0.017$). However, PRS Univariate degraded substantially from real ($0.638 \pm 0.023$) to synthetic ($0.506 \pm 0.020$), indicating that linear PRS is more sensitive to the subtle allele frequency shifts introduced by the generative process. Of note, BC uses 1{,}024 SNPs with different clumping parameters and achieved the lowest VAE reconstruction accuracy (73.1\%), likely contributing to this degradation.

\smallskip
\noindent The External PRS baseline (dashed lines in Figure~\ref{fig:downstream}) achieved AUCs comparable to PRS Univariate on real data, confirming that the selected SNPs carry genuine disease signal.

\smallskip
\noindent Overall, models trained on synthetic genotypes achieved AUCs consistently close to those obtained when training on real data, indicating that the generated cohorts retain usable disease-predictive signal for downstream risk prediction under the TSTR protocol. Of note, nonlinear models (XGBoost) consistently preserved more predictive signal from synthetic data than linear PRS, suggesting that interaction patterns remain intact even when marginal allele frequencies shift. This result suggests that conditional synthetic genomic datasets may be best suited for training nonlinear classifiers rather than constructing traditional additive models. Synthetic data also match or outperform the VAE-reconstructed data, because the VAE is unconditional and optimises reconstruction fidelity, whereas the diffusion model is explicitly conditioned on disease status and can steer samples toward phenotype-aligned genotype patterns.

\smallskip
\noindent Finally, in Table~\ref{tab:gwas_prs} we report the comparison of SNPgen against genome-wide PRS methods that use substantially more variants. The genome-wide comparison confirms that 1{,}024--2{,}048 trait-associated SNPs capture substantial polygenic signal. Where genome-wide methods retain an advantage (BC, T2D), the additional signal resides in the polygenic tail---aggregate weak effects that a targeted representation trades for scalability. This trade-off favours traits with concentrated genetic architectures, as the T1D results demonstrate.

\begin{table}[!b]
\caption{Genome-wide PRS baselines (PRS-only AUC).\label{tab:gwas_prs}}%
\begin{tabular*}{\textwidth}{@{\extracolsep\fill}lccc@{\extracolsep\fill}}
\toprule
Trait & PRSice-2 & LDpred2 & SNPgen \\
      & (best config) & (auto) & (best XGB) \\
\midrule
CAD   & 0.606 & 0.539 & 0.594 \\
BC    & 0.636 & 0.661 & 0.598 \\
T1D   & 0.628 & 0.559 & 0.671 \\
T2D   & 0.599 & 0.628 & 0.593 \\
\botrule
\end{tabular*}
\begin{tablenotes}%
\item PRSice-2: best of 10 clumping configurations per trait. LDpred2: HapMap3+ reference (${\sim}$1.2M SNPs). SNPgen: best AUC across Synthetic and Syn Augmented conditions with XGBoost or XGBoost Balanced.
\end{tablenotes}
\end{table}

\begin{figure}[!b]
\centering
\includegraphics[width=\textwidth]{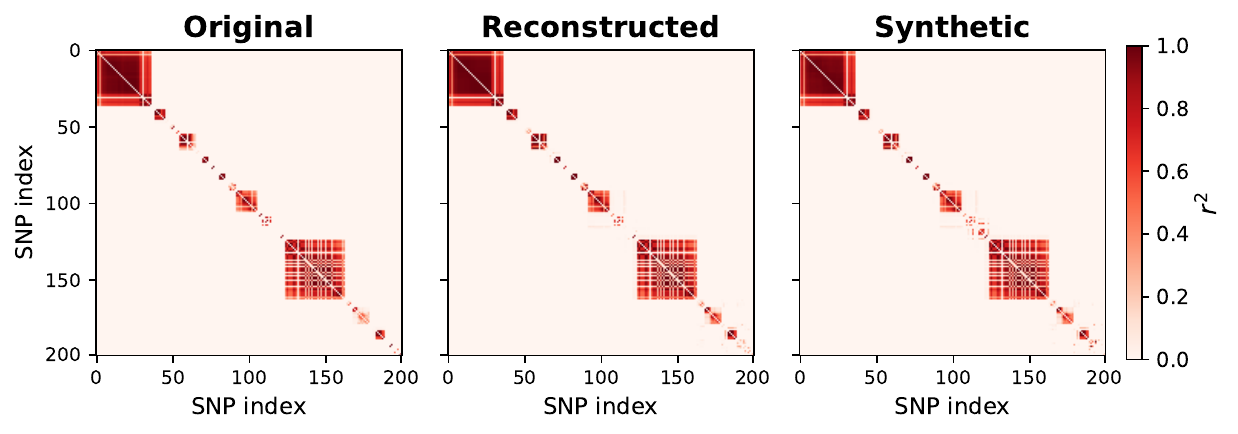}
\caption{Pairwise LD ($r^2$) for T2D (first 200 SNPs). Left: original. Centre: reconstructed. Right: synthetic. Block-diagonal structure is preserved.}\label{fig:ld}
\end{figure}

\begin{table}[!h]
\caption{Privacy evaluation summary. IMR: Identical Match Rate ($0\%$ = no copies). NNDR: Nearest Neighbour Distance Ratio (higher = more private). MI: Membership Inference AUC (${\approx}0.5$ = no leakage). NNAA: Nearest Neighbour Adversarial Accuracy (${\approx}0.5$ = no leakage). DCR$_{<5\%}$: fraction of synthetic samples closer to training than the 5th percentile of holdout--training distances (lower = more private). MAF~$r$: allele frequency correlation (higher = better population fidelity).\label{tab:privacy}}
\begin{tabular*}{\textwidth}{@{\extracolsep{\fill}}llcccccc@{\extracolsep{\fill}}}
\toprule%
Trait & Condition & IMR (\%) & NNDR & MI AUC & NNAA & DCR$_{<5\%}$ (\%) & MAF $r$ \\
\midrule
\multirow{2}{*}{CAD}
 & Reconstructed & 0.0 & 0.263 & 1.000 & 0.468 & 100.0 & 0.989 \\
 & Synthetic     & 0.0 & 0.983 & 0.500 & 0.537 & 3.5  & 0.988 \\
\midrule
\multirow{2}{*}{BC}
 & Reconstructed & 0.0 & 0.984 & 0.856 & 0.507 & 73.7 & 0.950 \\
 & Synthetic     & 0.0 & 0.990 & 0.501 & 0.507 & 46.3 & 0.951 \\
\midrule
\multirow{2}{*}{T1D}
 & Reconstructed & 0.0 & 0.247 & 0.979 & 0.460 & 86.3 & 1.000 \\
 & Synthetic     & 0.0 & 0.934 & 0.503 & 0.572 & 2.3  & 0.999 \\
\midrule
\multirow{2}{*}{T2D}
 & Reconstructed & 0.0 & 0.288 & 1.000 & 0.456 & 100.0 & 0.988 \\
 & Synthetic     & 0.0 & 0.983 & 0.499 & 0.522 & 4.7   & 0.987 \\
\botrule
\end{tabular*}
\end{table}

\subsection{Privacy analysis}\label{sec:results_privacy}

Table~\ref{tab:privacy} summarizes privacy metrics across the four UK Biobank traits. Synthetic data showed strong privacy preservation: zero exact matches (IMR~$= 0\%$), high NNDR ($\geq 0.934$), MI~AUC indistinguishable from random ($\approx 0.50$), and NNAA close to 0.5 ($\leq 0.572$). MAF correlation remained high ($r \geq 0.951$), confirming that population-level allele frequencies are preserved despite individual-level privacy.
In contrast, reconstructed data showed expected memorization (e.g., MI~AUC~$= 1.000$ for CAD and T2D), confirming that while the VAE provides a high-fidelity latent representation, the LDM generates genuinely novel samples with strong privacy guarantees.

\subsection{LD structure preservation}\label{sec:results_ld}

Figure~\ref{fig:ld} shows pairwise LD ($r^2$) matrices for T2D (representative trait). Both reconstructed and synthetic data preserved the block-diagonal LD structure of the original genotypes. MAF correlation values ($r = 0.95$--$0.99$ across traits, Table~\ref{tab:privacy}) provide additional quantitative confirmation of population-level genetic fidelity. LD comparison matrices for the remaining three traits and the simulated phenotype, as well as the results of the LD-decay evaluation, are provided in Supplementary Section~\ref{sec:supp_ld}.

\section{Conclusions}\label{sec:conclusions}

We introduced \textbf{SNPgen}, a phenotype-supervised framework for generating task-ready synthetic genotypes by combining GWAS-guided SNP selection with label-conditioned latent diffusion in a compact VAE latent space. Across the simulation and UK Biobank experiments (Figures~\ref{fig:downstream}--\ref{fig:ld}, Tables~\ref{tab:gwas_prs}--\ref{tab:privacy}, Supplementary Table~\ref{tab:performance}), SNPgen produces synthetic cohorts that retain clinically relevant signal while preserving key population-genetic structure, enabling \textbf{train-on-synthetic, test-on-real} model development and benchmarking when individual-level genotypes cannot be shared.

\smallskip
\noindent In a controlled simulated-trait setting with known causal effects, synthetic data recover marginal effect sizes more accurately than VAE reconstructions (higher beta correlation), providing direct evidence that phenotype conditioning improves genotype--phenotype association fidelity rather than merely reproducing LD or allele-frequency statistics. On real traits, SNPgen shows that a \textbf{targeted panel of 1--2\,k GWAS-prioritized variants can still support useful downstream risk modelling}---particularly with nonlinear predictors---while naturally trading off some performance relative to genome-wide PRS in traits dominated by the polygenic tail. Overall, by coupling signal-concentrating variant selection with phenotype-aligned generation, SNPgen reduces the common gap between genomic realism and downstream utility that characterises many unconditional generators, \textbf{while mitigating direct re-identification and membership leakage} risks under the evaluated privacy attacks.

\smallskip
\noindent Some limitations should be acknowledged. In particular, SNPgen currently targets trait-specific panels (1{,}024--2{,}048 SNPs) and was evaluated primarily in a single-ancestry setting; extending to larger or hybrid panels, and multi-ancestry training/evaluation, are important next steps. Moreover, the current study focuses on binary phenotypes; extending conditioning to continuous traits and to additional covariates (e.g., age, sex, genetic PCs) would broaden applicability. Finally, while empirical privacy signals are strong, integrating formal privacy mechanisms and systematically mapping the privacy--utility trade-off---especially for minority classes---would further strengthen readiness for real data-release scenarios.

\section*{Author contributions}
A.L.\ conceived the study, developed the methodology and software, conducted all experiments, and wrote the original draft. M.C.M.\ co-designed the study and the experiments, supervised the work and contributed to writing, review \& editing. N.P.\ contributed to the simulation setup. F.I., M.M.\ and E.D.A.\ provided supervision and resources.

\section*{Conflict of interest}
The authors declare that they have no competing interests.

\section*{Funding}
This work has been supported by Fondazione Cariplo grant n. 2024-1044.

\section*{Data availability}
The code is available at \url{https://github.com/ht-diva/SNPgen}. The data used in this study are from UK Biobank (application 102297), available to approved researchers at \url{https://www.ukbiobank.ac.uk/}.

\section*{Acknowledgements}
This work was supported by MUR, grant Dipartimento di Eccellenza 2023-2027.\\
Large language models were used as an aid to correct written text and as an aid for code writing.

\bibliographystyle{plainnat-etal}
\bibliography{reference}

\clearpage
\appendix
\beginsupplement

\input{supplementary}

\end{document}

%% file: supplementary.tex
\begin{center}
{\LARGE\bfseries Supplementary Materials}
\end{center}
\vspace{1em}

\refstepcounter{suppsec}\label{sec:supp_phenotype}
\section*{\thesuppsec.\quad Phenotype Definitions}

\paragraph{Ancestry selection.}
Participants were filtered on self-reported ethnicity (UK Biobank field~21000,
code~1), which encompasses White British~[1001], Irish~[1002], and any other
White background~[1003], yielding 458{,}724 individuals after intersection with
available genotype data.

\paragraph{Phenotype definitions.}
Phenotype definitions were derived from hospital episode statistics (HES),
self-reported diagnoses (UK Biobank field~20002), and operative procedure codes.

\begin{itemize}
\item \textbf{Coronary Artery Disease (CAD):} ICD-10 codes I21, I22, I23, I24.1,
  I25.2 (fields~41270, 40001, 40002); ICD-9
  codes 410, 411, 412, 429.79 (field~41271); OPCS-4 procedure codes K40.1--4, K41.1--4,
  K45.1--5, K49.1--2, K49.8--9, K50.2, K75.1--4, K75.8--9 (field~41272);
  self-reported non-cancer illness field~20002 code~1075 (heart attack /
  myocardial infarction); self-reported operation field~20004 codes~1070
  (coronary angioplasty), 1095 (coronary artery bypass grafts); or
  field~6150 code~1 (heart attack diagnosed by doctor).
\item \textbf{Breast Cancer (BC):} ICD-10 codes C50, C500--C506, C508, C509
  (fields~40006, 40001, 40002); ICD-9
  codes 174, 1749 (field~40013); or self-reported cancer field~20001 code~1002 (breast cancer).
  Female participants only.
\item \textbf{Type~1 Diabetes (T1D):} ICD-10 codes E100--E109, O240
  (fields~41270, 40001, 40002); ICD-9 codes 250.x1, 250.x3 (field~41271); self-reported non-cancer
  illness field~20002 code~1222 (type 1 diabetes).
\item \textbf{Type~2 Diabetes (T2D):} ICD-10 codes E11, E110--E119
  (fields~41270, 40001, 40002); ICD-9 codes 250.x0, 250.x2 (field~41271); self-reported non-cancer
  illness field~20002 code~1223 (type 2 diabetes).
\end{itemize}

\begin{table}[H]
\caption{Cohort demographics by trait. Sample prevalence is the case fraction in the UK Biobank cohort; population prevalence is used for liability-scale $R^2$ conversion.\label{tab:cohort}}
\begin{tabular*}{\textwidth}{@{\extracolsep\fill}lrrrcc@{\extracolsep\fill}}
\toprule
Trait & $n$ & Cases & Controls & Sample prev.\ (\%) & Pop.\ prev.\ (\%) \\
\midrule
BC  & 248{,}987 & 19{,}634 & 229{,}353 & 7.9 & 12.0 \\
CAD & 458{,}724 & 35{,}306 & 423{,}418 & 7.7 & 6.0 \\
T1D & 458{,}724 &  4{,}593 & 454{,}131 & 1.0 & 0.4 \\
T2D & 458{,}724 & 38{,}541 & 420{,}183 & 8.4 & 8.0 \\
\botrule
\end{tabular*}
\begin{tablenotes}
\item BC restricted to female participants. Population prevalence values are used by LDpred2-auto for liability-scale heritability estimation.
\end{tablenotes}
\end{table}

\refstepcounter{suppsec}\label{sec:supp_gwas}
\section*{\thesuppsec.\quad GWAS Summary Statistics Sources}

External GWAS summary statistics (not derived from UK Biobank) were used for
SNP selection via clumping and thresholding, and as input for genome-wide PRS
methods (PRSice-2 and LDpred2-auto). All studies used European-ancestry cohorts.

\begin{itemize}
\item \textbf{CAD:} CARDIoGRAMplusC4D 1000 Genomes-based meta-analysis
  \citep{nikpayComprehensive1000Genomes2015}. GWAS Catalog accession: GCST003116.
\item \textbf{BC:} Breast Cancer Association Consortium (BCAC) GWAS
  \citep{zhangGenomewideAssociationStudy2020}. Summary statistics from
  \url{https://www.ccge.medschl.cam.ac.uk/breast-cancer-association-consortium-bcac/data-data-access/summary-results/gwas-summary-associations}.
\item \textbf{T1D:} Multi-ancestry T1D GWAS \citep{michalekMultiancestryGenomewideAssociation2024}. GWAS Catalog
  accession: GCST90432066.
\item \textbf{T2D:} DIAGRAM Consortium 1000 Genomes meta-analysis Stage~1
  \citep{scottExpandedGenomeWideAssociation2017}. GWAS Catalog accession: GCST004773.
\end{itemize}

\paragraph{SNP selection via clumping and thresholding.}
For each trait, LD clumping was performed with PLINK2~\citep{changSecondgenerationPLINKRising2015}
on the corresponding external GWAS summary statistics.  From the clumped set,
the top $L$ SNPs ranked by $p$-value were retained as the trait-specific panel.
GWAS effect-allele betas were sign-flipped where necessary to match the
genotype encoding.  Table~\ref{tab:snp_selection} summarises the per-trait
clumping parameters.

\begin{table}[H]
\caption{Per-trait SNP selection parameters.  BC required a looser clumping
configuration due to sparser GWAS signal in the available summary
statistics.\label{tab:snp_selection}}
\begin{tabular*}{\textwidth}{@{\extracolsep\fill}lcccc@{\extracolsep\fill}}
\toprule
Trait & GWAS source & Clumping window & $r^2$ threshold & Final $L$ \\
\midrule
CAD & \citet{nikpayComprehensive1000Genomes2015} & 10\,kb & 0.5 & 2{,}048 \\
BC  & \citet{zhangGenomewideAssociationStudy2020} & 500\,kb & 0.1 & 1{,}024 \\
T1D & \citet{michalekMultiancestryGenomewideAssociation2024} & 10\,kb & 0.5 & 2{,}048 \\
T2D & \citet{scottExpandedGenomeWideAssociation2017} & 10\,kb & 0.5 & 2{,}048 \\
\botrule
\end{tabular*}
\end{table}

\refstepcounter{suppsec}\label{sec:supp_architecture}
\section*{\thesuppsec.\quad Detailed Architecture Specifications}

\refstepcounter{suppsubsec}\label{sec:supp_vae}
\subsection*{\thesuppsubsec\quad Stage~1: VAE Encoder and Decoder}

The variational autoencoder (VAE)~\citep{kingmaAutoEncodingVariationalBayes2013} consists of an encoder and decoder with the following specifications.

\paragraph{Encoder.}
An input projection layer (Conv1d, $3 \to 32$ channels, kernel size~3) is followed
by five resolution levels with channel progression $[32, 64, 64, 128, 128]$, each
containing two residual blocks~\citep{heDeepResidualLearning2015}. Downsampling uses stride-2 convolutions at four
levels. A middle block with self-attention processes the deepest features at spatial
resolution $L/16$. The encoder outputs mean $\boldsymbol{\mu}$ and log-variance
$\log \boldsymbol{\sigma}^2$.

For $L{=}2048$ (CAD, T1D, T2D), the latent dimension is~128; for $L{=}1024$ (BC),
the latent dimension is~64. Both use $z_\text{channels}{=}1$.

\paragraph{Decoder.}
The decoder mirrors the encoder architecture, transposing downsampling into
upsampling stages (stride-2 transposed convolutions). The output layer produces
3-channel logits over the genotype alphabet $\{0, 1, 2\}$. Discrete genotypes are
recovered via $\arg\max$ over the channel dimension.

\paragraph{Discriminator.}
The Wasserstein discriminator (WDiscriminator) is a 6-layer 1D CNN with 128 hidden
channels per layer, spectral normalization~\citep{miyatoSpectralNormalizationGenerative2018}, instance normalization, and
LeakyReLU($0.2$) activations. It is trained with hinge loss and activated after
iteration~20{,}001 to allow the VAE reconstruction to stabilize before introducing
the discriminator signal.

\refstepcounter{suppsubsec}\label{sec:supp_ldm}
\subsection*{\thesuppsubsec\quad Stage~2: Latent Diffusion Model (UNet)}

\paragraph{Denoiser.}
The 1D UNet~\citep{ronnebergerUNetConvolutionalNetworks2015} is adapted from the Stable Diffusion UNet~\citep{rombachHighResolutionImageSynthesis2022},
replacing all 2D operations with 1D counterparts.
It uses \texttt{model\_channels}~$= 64$ and channel multipliers $[1, 2, 4]$,
yielding feature maps of $[64, 128, 256]$ channels across three resolution levels.
Each scale contains two residual blocks and spatial transformer attention
blocks~\citep{vaswaniAttentionAllYou2017} with 32-channel attention heads and transformer
depth~1. The xFormers~\citep{xFormers2022} memory-efficient attention
backend is used.

\paragraph{Conditioning.}
Binary phenotype labels are embedded via a learned ClassEmbedder with embedding
dimension~128 (for CAD, T1D, T2D) or~64 (for BC). Embeddings are injected through
cross-attention in the UNet transformer blocks. Classifier-free
guidance~\citep{hoClassifierFreeDiffusionGuidance2022} uses an unconditional dropout rate of~0.2 during training
and a guidance scale of~5.0 at inference.

\refstepcounter{suppsec}\label{sec:supp_hyperparams}
\section*{\thesuppsec.\quad Training Hyperparameters and Evaluation Grid}

\refstepcounter{suppsubsec}\label{sec:supp_training}
\subsection*{\thesuppsubsec\quad SNPgen Training}

\paragraph{VAE.}
Adam optimizer~\citep{kingmaAdamMethodStochastic2014}, learning rate $4.5 \times 10^{-6}$ (both
generator and discriminator), batch size~256, FP16 mixed precision, exponential
moving average (EMA) with decay~0.9999. Training lasted
400 epochs across all traits, selecting the checkpoint with the highest validation
reconstruction accuracy (per-SNP argmax agreement).

\paragraph{LDM.}
Adam optimizer, learning rate $10^{-4}$, cosine learning rate schedule with
5-epoch linear warmup, batch size~1024, FP16 mixed precision, EMA with
decay~0.9999. Training lasted 500 epochs across all traits, selecting the
checkpoint with the lowest validation loss.

\paragraph{Generation.}
Synthetic genotypes were sampled using the Euler EDM sampler~\citep{karrasElucidatingDesignSpace2022}
with 50 denoising steps and a classifier-free guidance scale of~5.0.
Latent vectors were decoded to discrete genotypes via the frozen VAE decoder
followed by $\arg\max$ over the channel dimension.

\paragraph{Hardware.}
All experiments were conducted on a single NVIDIA Tesla V100 GPU (32\,GB).

\refstepcounter{suppsubsec}\label{sec:supp_downstream_models}
\subsection*{\thesuppsubsec\quad Downstream Evaluation Models}

\paragraph{XGBoost.}
Gradient boosting via the \texttt{xgboost} Python package~\citep{chenXGBoostScalableTree2016} (GPU-accelerated).
Grid search over 100 combinations using 5-fold stratified inner CV:
\begin{itemize}
\item \texttt{max\_depth} $\in \{1, 2, 3, 6, 20\}$
\item \texttt{n\_estimators} $\in \{100, 500, 700, 800, 1000\}$
\item \texttt{learning\_rate} $\in \{0.01, 0.03, 0.05, 0.1\}$
\end{itemize}
For binary classification, \texttt{scale\_pos\_weight} is set to the exact ratio
$n_\text{controls}/n_\text{cases}$ computed from the training data.
Training uses the full (imbalanced) training set.

\paragraph{XGBoost (Balanced).}
Same model class, but trained on a \emph{class-balanced} subset obtained by
downsampling the majority class to match the minority class count. The
hyperparameter grid is coarser (36 combinations):
\begin{itemize}
\item \texttt{max\_depth} $\in \{1, 3, 6\}$
\item \texttt{n\_estimators} $\in \{100, 500, 1{,}000\}$
\item \texttt{learning\_rate} $\in \{0.01, 0.1\}$
\item \texttt{scale\_pos\_weight} $\in \{1, 11\}$ (binary classification only)
\end{itemize}

\paragraph{PRS Univariate.}
For each SNP independently, a univariate logistic regression is fitted with no
regularization, using the \texttt{glum} package. This is conceptually equivalent
to performing a GWAS on the training SNPs: the per-SNP effect sizes
$\hat{\beta}_i$ are estimated from the training cohort genotype--phenotype
association, rather than from an external GWAS study. The polygenic risk score
$\text{PRS} = \sum_i \hat{\beta}_i \cdot g_i$ is computed as a dot product of
betas and genotype dosages.

\paragraph{GWAS PRS baseline.}
Uses pre-existing effect sizes from the external GWAS summary statistics listed
in Supplementary Section~\ref{sec:supp_gwas} (not learned from training data). The PRS is computed as
$\text{PRS} = \sum_i \beta_i^\text{GWAS} \cdot g_i$ using fixed betas aligned
to the genotype encoding at dataset preparation time. If the Spearman
correlation between PRS and target is negative (sign convention mismatch), the
PRS is flipped.

\refstepcounter{suppsec}\label{sec:supp_downstream}
\section*{\thesuppsec.\quad Full Downstream Risk Prediction Results}

Table~\ref{tab:performance} reports the complete numerical results for the
downstream risk prediction evaluation summarized in
Figure~\ref{fig:downstream} of the main text.

\begin{table}[H]
\caption{Downstream risk prediction performance (ROC-AUC $\pm$ 95\% CI). \textbf{Bold}: best result per trait and model (excluding Real). \underline{Underline}: second best.\label{tab:performance}}
\begin{tabular*}{\textwidth}{@{\extracolsep{\fill}}llccc@{\extracolsep{\fill}}}
\toprule%
Trait & Condition & XGBoost & XGBoost (Balanced) & PRS Univariate \\
\midrule
\multirow{4}{*}{Simulated}
 & \textit{Real}    & \textit{0.967 $\pm$ 0.002} & \textit{0.975 $\pm$ 0.002} & \textit{0.903 $\pm$ 0.006} \\
 & Reconstructed    & \textbf{0.960 $\pm$ 0.003} & \textbf{0.967 $\pm$ 0.003} & \textbf{0.898 $\pm$ 0.007} \\
 & Synthetic        & 0.902 $\pm$ 0.012 & 0.897 $\pm$ 0.006 & 0.849 $\pm$ 0.008 \\
 & Syn Augmented    & \underline{0.905 $\pm$ 0.011} & \underline{0.898 $\pm$ 0.006} & \underline{0.849 $\pm$ 0.008} \\
\midrule
\multirow{4}{*}{CAD}
 & \textit{Real}    & \textit{0.592 $\pm$ 0.031} & \textit{0.601 $\pm$ 0.009} & \textit{0.575 $\pm$ 0.014} \\
 & Reconstructed    & 0.559 $\pm$ 0.036 & 0.578 $\pm$ 0.014 & 0.566 $\pm$ 0.019 \\
 & Synthetic        & \underline{0.589 $\pm$ 0.009} & \underline{0.590 $\pm$ 0.009} & \textbf{0.568 $\pm$ 0.017} \\
 & Syn Augmented    & \textbf{0.594 $\pm$ 0.011} & \textbf{0.592 $\pm$ 0.011} & \underline{0.568 $\pm$ 0.015} \\
\midrule
\multirow{4}{*}{BC}
 & \textit{Real}    & \textit{0.624 $\pm$ 0.026} & \textit{0.627 $\pm$ 0.018} & \textit{0.638 $\pm$ 0.023} \\
 & Reconstructed    & 0.564 $\pm$ 0.043 & 0.564 $\pm$ 0.009 & 0.504 $\pm$ 0.018 \\
 & Synthetic        & \underline{0.591 $\pm$ 0.023} & \underline{0.588 $\pm$ 0.015} & \underline{0.506 $\pm$ 0.020} \\
 & Syn Augmented    & \textbf{0.598 $\pm$ 0.017} & \textbf{0.597 $\pm$ 0.017} & \textbf{0.547 $\pm$ 0.018} \\
\midrule
\multirow{4}{*}{T1D}
 & \textit{Real}    & \textit{0.655 $\pm$ 0.038} & \textit{0.668 $\pm$ 0.033} & \textit{0.649 $\pm$ 0.038} \\
 & Reconstructed    & 0.641 $\pm$ 0.036 & 0.657 $\pm$ 0.044 & \textbf{0.649 $\pm$ 0.038} \\
 & Synthetic        & \underline{0.659 $\pm$ 0.034} & \underline{0.670 $\pm$ 0.022} & \underline{0.647 $\pm$ 0.034} \\
 & Syn Augmented    & \textbf{0.671 $\pm$ 0.026} & \textbf{0.671 $\pm$ 0.024} & 0.647 $\pm$ 0.034 \\
\midrule
\multirow{4}{*}{T2D}
 & \textit{Real}    & \textit{0.589 $\pm$ 0.041} & \textit{0.607 $\pm$ 0.016} & \textit{0.587 $\pm$ 0.011} \\
 & Reconstructed    & 0.585 $\pm$ 0.032 & 0.588 $\pm$ 0.024 & 0.560 $\pm$ 0.013 \\
 & Synthetic        & \underline{0.587 $\pm$ 0.019} & \underline{0.591 $\pm$ 0.015} & \underline{0.567 $\pm$ 0.011} \\
 & Syn Augmented    & \textbf{0.592 $\pm$ 0.016} & \textbf{0.593 $\pm$ 0.017} & \textbf{0.576 $\pm$ 0.011} \\
\botrule
\end{tabular*}
\begin{tablenotes}%
\item Cohort: Simulated $n{=}458{,}724$ (150{,}901 cases, 32.9\% prevalence); CAD $n{=}458{,}724$ (35{,}306 cases); BC $n{=}248{,}987$ (19{,}634 cases, female only); T1D $n{=}458{,}724$ (4{,}593 cases); T2D $n{=}458{,}724$ (38{,}541 cases).
\item External PRS AUC baselines (fixed external betas): Simulated ${\approx}$1.000 (imposed betas); CAD 0.574, BC 0.633, T1D 0.650, T2D 0.584.
\end{tablenotes}
\end{table}

\refstepcounter{suppsec}\label{sec:supp_prsice}
\section*{\thesuppsec.\quad PRSice-2 Full Results}

PRSice-2~\citep{choiPRSice2PolygenicRisk2019} was evaluated with 10 clumping configurations
(window $\in \{10, 50, 100, 250, 500\}$\,kb $\times$ $r^2 \in \{0.1, 0.5\}$)
per trait. Tables~\ref{tab:prsice_bc}--\ref{tab:prsice_t2d} report all
configurations. The best configuration per trait (by AUC, PRS only) is shown
in \textbf{bold}. All AUC values were obtained from the held-out test set.

\begin{table}[H]
\caption{PRSice-2 results for Coronary Artery Disease (CAD). $n = 458{,}724$
($35{,}306$ cases). \textbf{Bold}: best AUC.\label{tab:prsice_cad}}
\begin{tabular*}{\textwidth}{@{\extracolsep\fill}lrccccc@{\extracolsep\fill}}
\toprule
Config & N SNPs & AUC (PRS Only) & AUC (Age+Sex) & AUC (Age+Sex+PRS) & $\Delta$AUC & $R^2_\text{liab}$ \\
\midrule
10\,kb / 0.1    &  19{,}867 & 0.587 & 0.713 & 0.709 & $-$0.004 & 0.052 \\
10\,kb / 0.5    &  23{,}517 & 0.587 & 0.713 & 0.708 & $-$0.004 & 0.052 \\
50\,kb / 0.1    &   3{,}878 & 0.593 & 0.713 & 0.710 & $-$0.003 & 0.060 \\
50\,kb / 0.5    &   5{,}706 & 0.596 & 0.713 & 0.711 & $-$0.002 & 0.064 \\
100\,kb / 0.1   &   1{,}724 & 0.596 & 0.713 & 0.711 & $-$0.002 & 0.065 \\
100\,kb / 0.5   &   4{,}954 & 0.602 & 0.713 & 0.713 & $+$0.000 & 0.072 \\
250\,kb / 0.1   &       641 & 0.600 & 0.713 & 0.712 & $-$0.001 & 0.071 \\
250\,kb / 0.5   &   3{,}955 & 0.605 & 0.713 & 0.714 & $+$0.001 & 0.077 \\
500\,kb / 0.1   &       606 & 0.601 & 0.713 & 0.713 & $+$0.000 & 0.073 \\
\textbf{500\,kb / 0.5} & \textbf{3{,}881} & \textbf{0.606} & \textbf{0.713} & \textbf{0.714} & \textbf{$+$0.001} & \textbf{0.079} \\
\botrule
\end{tabular*}
\end{table}

\begin{table}[H]
\caption{PRSice-2 results for Breast Cancer (BC). $n = 248{,}987$
($19{,}634$ cases). \textbf{Bold}: best AUC.\label{tab:prsice_bc}}
\begin{tabular*}{\textwidth}{@{\extracolsep\fill}lrccccc@{\extracolsep\fill}}
\toprule
Config & N SNPs & AUC (PRS Only) & AUC (Age+Sex) & AUC (Age+Sex+PRS) & $\Delta$AUC & $R^2_\text{liab}$ \\
\midrule
10\,kb / 0.1    & 110{,}465 & 0.595 & 0.581 & 0.626 & +0.045 & 0.077 \\
10\,kb / 0.5    & 109{,}954 & 0.600 & 0.581 & 0.629 & +0.048 & 0.084 \\
50\,kb / 0.1    &       617 & 0.604 & 0.581 & 0.632 & +0.051 & 0.089 \\
50\,kb / 0.5    &  15{,}193 & 0.613 & 0.581 & 0.639 & +0.058 & 0.107 \\
100\,kb / 0.1   &       458 & 0.613 & 0.581 & 0.639 & +0.058 & 0.108 \\
100\,kb / 0.5   &   9{,}493 & 0.623 & 0.581 & 0.647 & +0.066 & 0.127 \\
250\,kb / 0.1   &   1{,}940 & 0.624 & 0.581 & 0.648 & +0.067 & 0.130 \\
250\,kb / 0.5   &   7{,}924 & 0.633 & 0.581 & 0.656 & +0.075 & 0.149 \\
500\,kb / 0.1   &   1{,}121 & 0.633 & 0.581 & 0.656 & +0.075 & 0.151 \\
\textbf{500\,kb / 0.5} & \textbf{6{,}786} & \textbf{0.636} & \textbf{0.581} & \textbf{0.658} & \textbf{+0.077} & \textbf{0.157} \\
\botrule
\end{tabular*}
\end{table}

\begin{table}[H]
\caption{PRSice-2 results for Type~1 Diabetes (T1D). $n = 458{,}724$
($4{,}593$ cases). \textbf{Bold}: best AUC. Note: the
\texttt{kb\_500\_r2\_0.5} configuration was not available for this
trait.\label{tab:prsice_t1d}}
\begin{tabular*}{\textwidth}{@{\extracolsep\fill}lrccccc@{\extracolsep\fill}}
\toprule
Config & N SNPs & AUC (PRS Only) & AUC (Age+Sex) & AUC (Age+Sex+PRS) & $\Delta$AUC & $R^2_\text{liab}$ \\
\midrule
10\,kb / 0.1    &   1{,}040 & 0.616 & 0.586 & 0.643 & +0.057 & 0.201 \\
10\,kb / 0.5    &   1{,}632 & 0.623 & 0.586 & 0.648 & +0.062 & 0.230 \\
50\,kb / 0.1    &       356 & 0.619 & 0.586 & 0.645 & +0.060 & 0.223 \\
50\,kb / 0.5    &       799 & 0.625 & 0.586 & 0.650 & +0.064 & 0.248 \\
100\,kb / 0.1   &       234 & 0.623 & 0.586 & 0.649 & +0.063 & 0.246 \\
100\,kb / 0.5   &       646 & 0.626 & 0.586 & 0.652 & +0.066 & 0.261 \\
\textbf{250\,kb / 0.1} & \textbf{137} & \textbf{0.628} & \textbf{0.586} & \textbf{0.653} & \textbf{+0.068} & \textbf{0.277} \\
250\,kb / 0.5   &       560 & 0.628 & 0.586 & 0.653 & +0.068 & 0.275 \\
500\,kb / 0.1   &       110 & 0.626 & 0.586 & 0.652 & +0.067 & 0.276 \\
\botrule
\end{tabular*}
\end{table}

\begin{table}[H]
\caption{PRSice-2 results for Type~2 Diabetes (T2D). $n = 458{,}724$
($38{,}541$ cases). \textbf{Bold}: best AUC.\label{tab:prsice_t2d}}
\begin{tabular*}{\textwidth}{@{\extracolsep\fill}lrccccc@{\extracolsep\fill}}
\toprule
Config & N SNPs & AUC (PRS Only) & AUC (Age+Sex) & AUC (Age+Sex+PRS) & $\Delta$AUC & $R^2_\text{liab}$ \\
\midrule
10\,kb / 0.1    & 133{,}010 & 0.590 & 0.619 & 0.639 & +0.020 & 0.056 \\
10\,kb / 0.5    & 156{,}002 & 0.593 & 0.619 & 0.641 & +0.022 & 0.060 \\
50\,kb / 0.1    &   1{,}383 & 0.589 & 0.619 & 0.638 & +0.019 & 0.056 \\
50\,kb / 0.5    &  20{,}591 & 0.597 & 0.619 & 0.643 & +0.024 & 0.064 \\
100\,kb / 0.1   &   1{,}137 & 0.589 & 0.619 & 0.638 & +0.019 & 0.056 \\
100\,kb / 0.5   &   5{,}007 & 0.597 & 0.619 & 0.643 & +0.024 & 0.066 \\
250\,kb / 0.1   &       298 & 0.593 & 0.619 & 0.641 & +0.021 & 0.060 \\
\textbf{250\,kb / 0.5} & \textbf{4{,}351} & \textbf{0.599} & \textbf{0.619} & \textbf{0.644} & \textbf{+0.025} & \textbf{0.069} \\
500\,kb / 0.1   &       281 & 0.594 & 0.619 & 0.641 & +0.022 & 0.061 \\
500\,kb / 0.5   &   2{,}631 & 0.599 & 0.619 & 0.644 & +0.025 & 0.069 \\
\botrule
\end{tabular*}
\end{table}

\refstepcounter{suppsec}\label{sec:supp_ldpred2}
\section*{\thesuppsec.\quad LDpred2-auto Results}

LDpred2-auto~\citep{priveLDpred2BetterFaster2021} was evaluated using the HapMap3+ LD
reference (${\sim}$1.2M variants), 30 chains with 500 burn-in and 500 sampling
iterations.

\begin{table}[H]
\caption{LDpred2-auto results. $\Delta$AUC: improvement over Age+Sex baseline.
$R^2_\text{liab}$: liability-scale $R^2$ increment of PRS.\label{tab:ldpred2}}
\resizebox{\textwidth}{!}{%
\begin{tabular}{lrrccccccccc}
\toprule
Trait & $n$ & Cases & OR/SD & AUC (PRS Only) & AUC (A+S) & AUC (A+S+PRS) & $\Delta$AUC & AUC (Full) & AUC (Full+PRS) & $R^2_\text{liab}$ & PR-AUC \\
\midrule
BC  & 248{,}987 & 19{,}634 & 1.822 & 0.661 & 0.581 & 0.679 & $+$0.098 & 0.582 & 0.679 & 0.221 & 0.147 \\
CAD & 458{,}724 & 35{,}306 & 1.165 & 0.539 & 0.713 & 0.705 & $-$0.007 & 0.708 & 0.708 & 0.012 & 0.089 \\
T1D & 458{,}724 &  4{,}593 & 1.235 & 0.559 & 0.586 & 0.599 & $+$0.013 & 0.584 & 0.602 & 0.046 & 0.013 \\
T2D & 458{,}724 & 38{,}541 & 1.602 & 0.628 & 0.619 & 0.665 & $+$0.046 & 0.621 & 0.671 & 0.115 & 0.134 \\
\botrule
\end{tabular}%
}
\begin{tablenotes}
\item A+S: Age+Sex covariates. Full: all covariates (age, sex, first 10 genetic principal components).
CAD: the negative $\Delta$AUC suggests LDpred2 PRS may be affected by GWAS--reference panel mismatch for this
trait; the Age+Sex baseline AUC is already 0.713, leaving little room for PRS-based improvement.
\end{tablenotes}
\end{table}

\refstepcounter{suppsec}\label{sec:supp_privacy_class}
\section*{\thesuppsec.\quad Class-Specific Privacy Evaluation}

Table~\ref{tab:privacy_class} reports privacy metrics stratified by phenotype
class (controls, label~0; cases, label~1).  For each class, all distances and
statistics are computed exclusively within that class: synthetic controls are
compared only to real training controls, and likewise for cases.  NNAA is
evaluated on up to 50{,}000 randomly sampled individuals per set within each
class.

Class-specific analysis reveals that minority-class samples (cases) are more
vulnerable to privacy leakage, particularly in the reconstructed condition.
For CAD and T2D, reconstructed cases show NNAA~$= 0.000$, indicating that the
VAE reconstruction of every case individual is closer to the original training
set than to the holdout---an expected consequence of the VAE's deterministic
encoding for high-fidelity reconstruction.  T1D reconstructed cases similarly
show NNAA~$= 0.001$.  The synthetic condition is substantially more private:
case-specific MI~AUC remains near~0.5 across all traits, and NNAA stays above
0.50.  However, case-specific DCR$_{<5\%}$ for synthetic data is elevated
compared to the overall values (e.g., T2D: 40.0\% for cases vs.\ 4.7\%
overall), reflecting the smaller effective sample size in the minority class
and the higher density of the case manifold.

\begin{table}[H]
\caption{Class-specific privacy metrics. All metrics are computed within each
class independently: controls (label~0) and cases (label~1).  NNAA computed
on up to 50{,}000 samples per set within each class.
Abbreviations as in Table~\ref{tab:privacy} of the main text.\label{tab:privacy_class}}
{\small
\begin{tabular*}{\textwidth}{@{\extracolsep{\fill}}lllccccc@{\extracolsep{\fill}}}
\toprule
Trait & Condition & Class & NNDR & MI AUC & NNAA & DCR$_{<5\%}$ (\%) & MAF $r$ \\
\midrule
\multirow{4}{*}{Simulated}
 & \multirow{2}{*}{Reconstructed} & Controls & 0.260 & 1.000 & 0.438 & 100.0 & 0.989 \\
 &                                & Cases    & 0.260 & 1.000 & 0.334 & 100.0 & 0.989 \\
 & \multirow{2}{*}{Synthetic}     & Controls & 0.982 & 0.502 & 0.583 & 5.3   & 0.979 \\
 &                                & Cases    & 0.983 & 0.502 & 0.546 & 34.5  & 0.984 \\
\midrule
\multirow{4}{*}{CAD}
 & \multirow{2}{*}{Reconstructed} & Controls & 0.262 & 1.000 & 0.464 & 100.0 & 0.989 \\
 &                                & Cases    & 0.249 & 1.000 & 0.000 & 100.0 & 0.989 \\
 & \multirow{2}{*}{Synthetic}     & Controls & 0.983 & 0.500 & 0.537 & 5.1   & 0.988 \\
 &                                & Cases    & 0.981 & 0.501 & 0.542 & 27.1  & 0.984 \\
\midrule
\multirow{4}{*}{BC}
 & \multirow{2}{*}{Reconstructed} & Controls & 0.984 & 0.858 & 0.507 & 95.2  & 0.950 \\
 &                                & Cases    & 0.975 & 0.902 & 0.503 & 100.0 & 0.950 \\
 & \multirow{2}{*}{Synthetic}     & Controls & 0.990 & 0.501 & 0.508 & 81.4  & 0.951 \\
 &                                & Cases    & 0.987 & 0.502 & 0.504 & 99.7  & 0.950 \\
\midrule
\multirow{4}{*}{T1D}
 & \multirow{2}{*}{Reconstructed} & Controls & 0.247 & 0.979 & 0.456 & 87.4  & 1.000 \\
 &                                & Cases    & 0.150 & 0.999 & 0.001 & 99.6  & 1.000 \\
 & \multirow{2}{*}{Synthetic}     & Controls & 0.934 & 0.503 & 0.572 & 2.4   & 0.998 \\
 &                                & Cases    & 0.912 & 0.482 & 0.605 & 5.8   & 0.984 \\
\midrule
\multirow{4}{*}{T2D}
 & \multirow{2}{*}{Reconstructed} & Controls & 0.287 & 1.000 & 0.453 & 100.0 & 0.988 \\
 &                                & Cases    & 0.274 & 1.000 & 0.000 & 100.0 & 0.988 \\
 & \multirow{2}{*}{Synthetic}     & Controls & 0.983 & 0.499 & 0.523 & 6.3   & 0.987 \\
 &                                & Cases    & 0.982 & 0.496 & 0.528 & 40.0  & 0.984 \\
\botrule
\end{tabular*}
}
\begin{tablenotes}
\item IMR is 0.0\% for all rows and is omitted for brevity.
\end{tablenotes}
\end{table}

\refstepcounter{suppsec}\label{sec:supp_ld}
\section*{\thesuppsec.\quad LD Structure Preservation for All Traits}

Pairwise LD ($r^2$, squared Pearson correlation) was computed for the original,
VAE-reconstructed, and LDM-generated synthetic datasets. The full LD matrices
(without restricting to a SNP subset) are shown below for all four real traits
and the simulated phenotype.

\begin{figure}[H]
\centering
\includegraphics[width=0.9\textwidth]{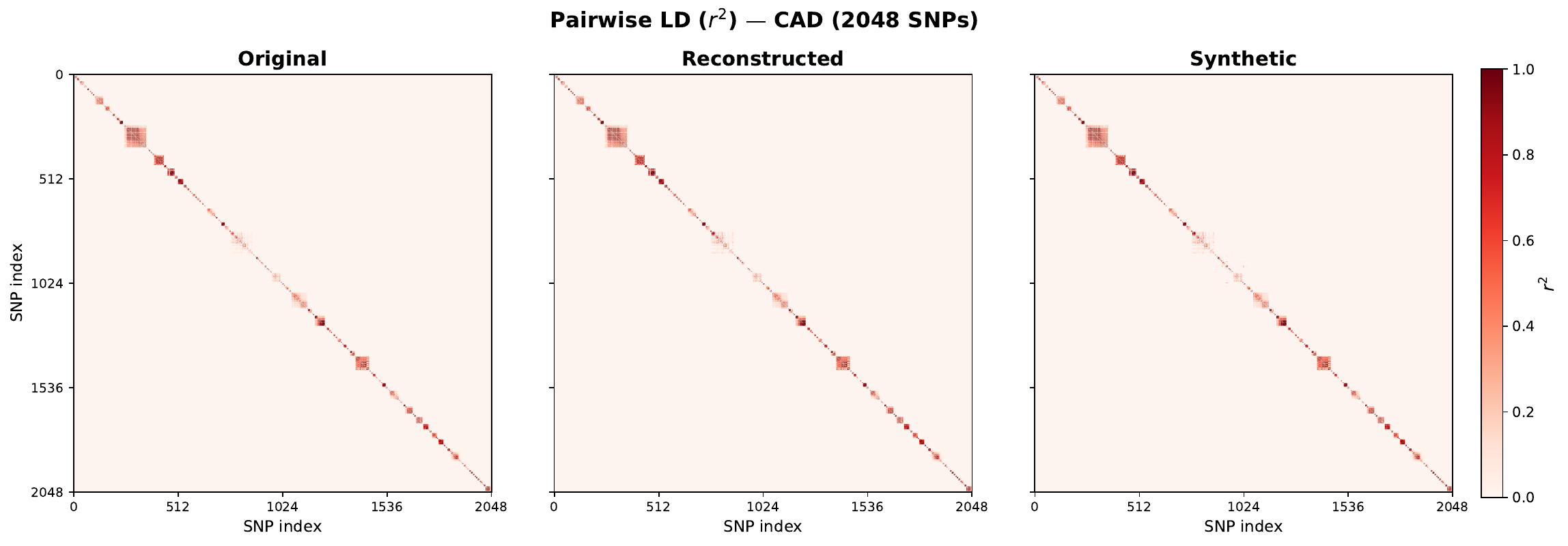}
\caption{Pairwise LD ($r^2$) for CAD (2{,}048 SNPs). Left: original.
Centre: VAE-reconstructed. Right: LDM-generated
synthetic.\label{fig:supp_ld_cad}}
\end{figure}

\begin{figure}[H]
\centering
\includegraphics[width=0.9\textwidth]{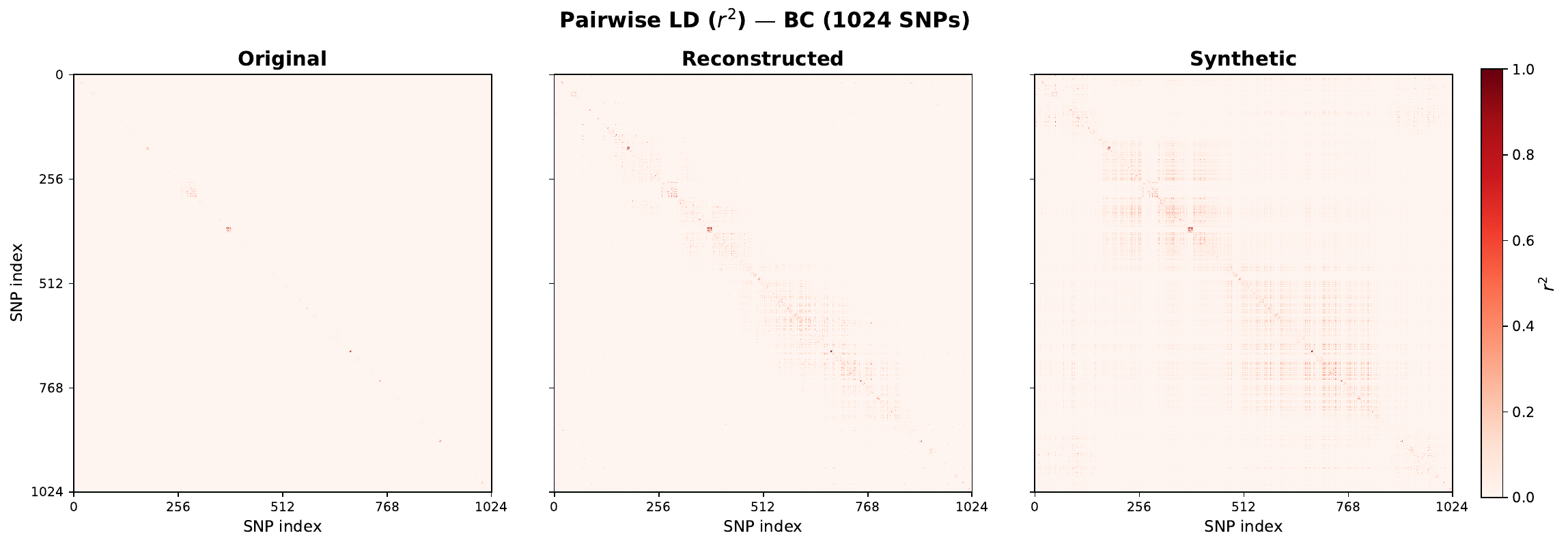}
\caption{Pairwise LD ($r^2$) for BC (1{,}024 SNPs). Left: original.
Centre: VAE-reconstructed. Right: LDM-generated
synthetic.\label{fig:supp_ld_bc}}
\end{figure}

\begin{figure}[H]
\centering
\includegraphics[width=0.9\textwidth]{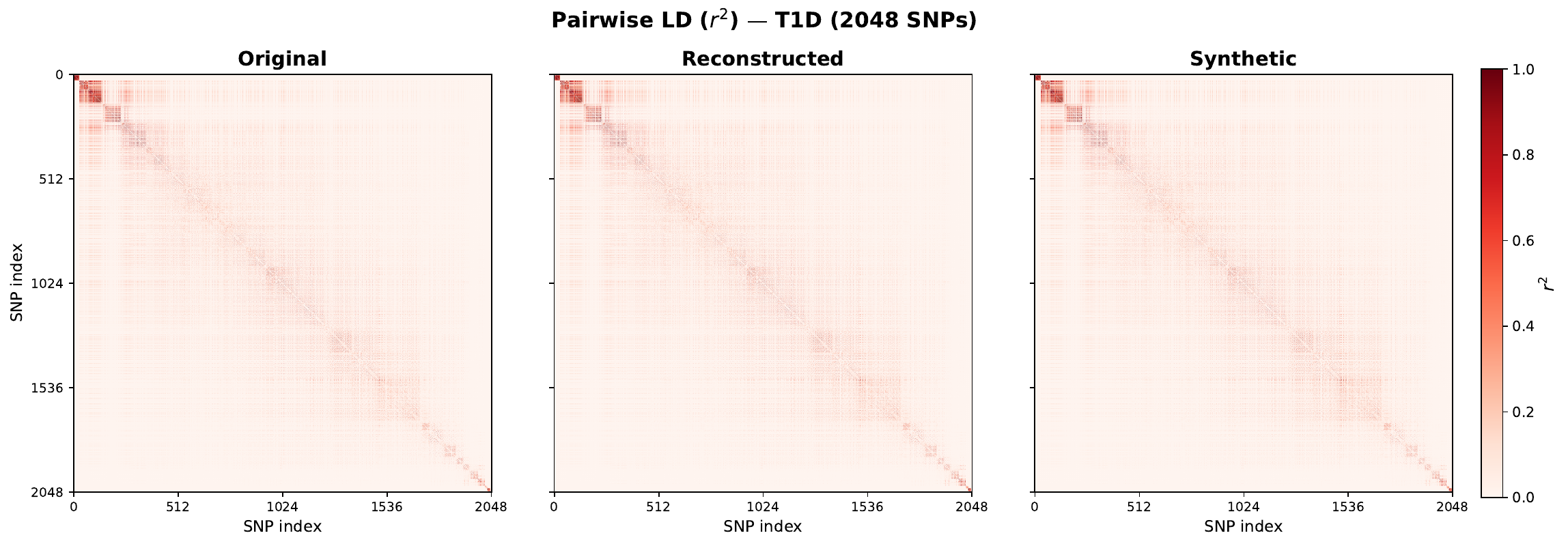}
\caption{Pairwise LD ($r^2$) for T1D (2{,}048 SNPs). Left: original.
Centre: VAE-reconstructed. Right: LDM-generated
synthetic.\label{fig:supp_ld_t1d}}
\end{figure}

\begin{figure}[H]
\centering
\includegraphics[width=0.9\textwidth]{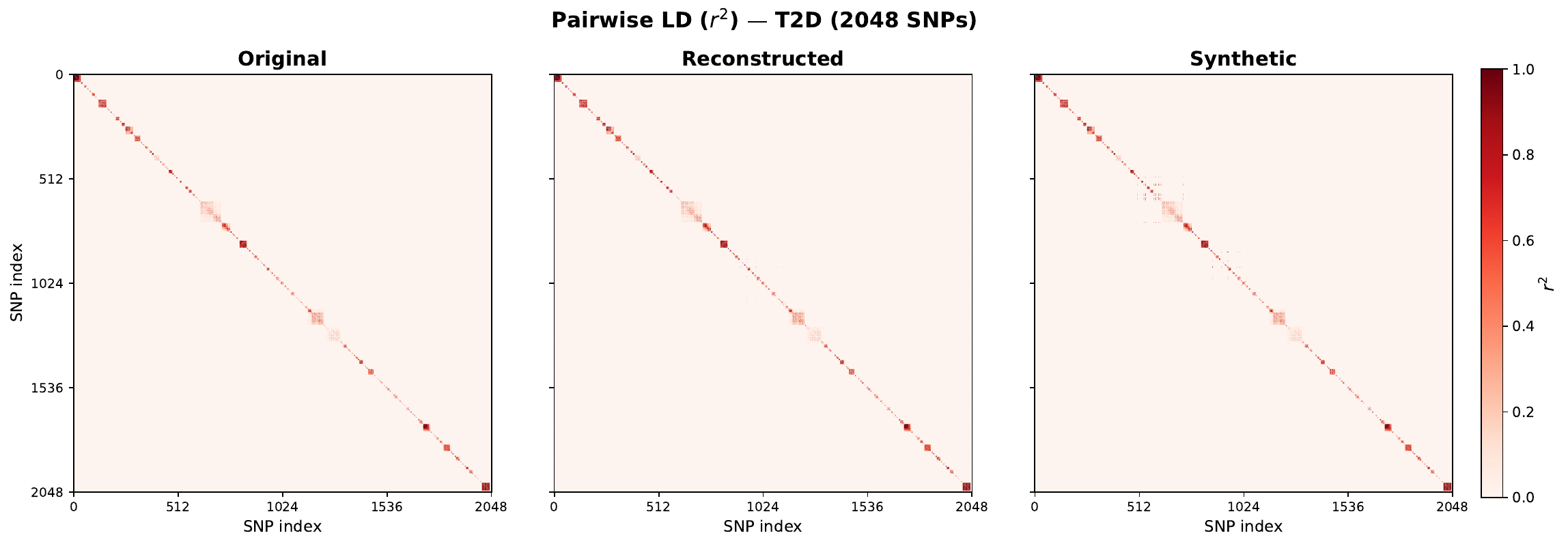}
\caption{Pairwise LD ($r^2$) for T2D (2{,}048 SNPs). This is the full
matrix corresponding to the 200-SNP subset shown in the main paper
(Figure~\ref{fig:ld}). Left: original. Centre: VAE-reconstructed. Right:
LDM-generated synthetic.\label{fig:supp_ld_t2d}}
\end{figure}

\begin{figure}[H]
\centering
\includegraphics[width=0.9\textwidth]{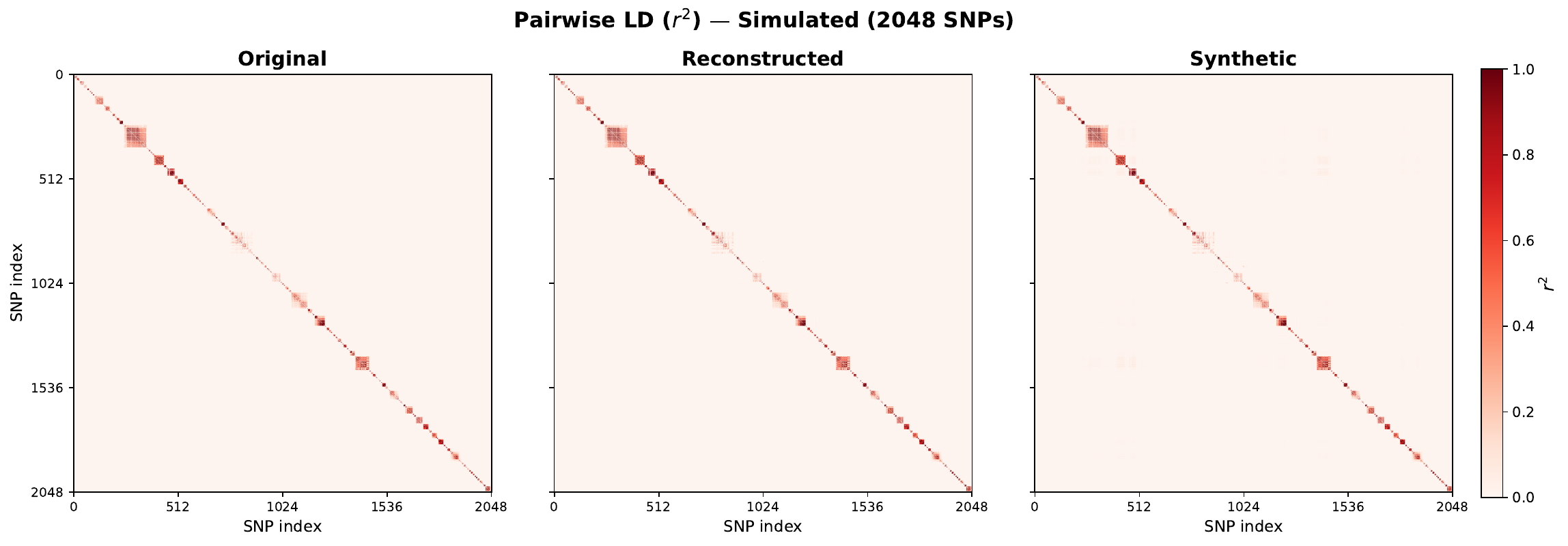}
\caption{Pairwise LD ($r^2$) for the simulated phenotype (2{,}048 SNPs, same
variant set as CAD). Left: original. Centre: VAE-reconstructed. Right:
LDM-generated synthetic.\label{fig:supp_ld_simulated}}
\end{figure}

\paragraph{LD decay with physical distance.}
To quantify LD preservation beyond the heatmap visualisation, we computed mean
$r^2$ as a function of physical distance (base pairs) for all same-chromosome
SNP pairs.  Because our GWAS-guided selection draws SNPs from across all~22
autosomes, only a fraction of all SNP pairs lie on the same chromosome
(6--7\% for CAD, T2D, BC, and Simulated; 94\% for T1D, whose selected SNPs
concentrate on chromosome~6).  Cross-chromosome pairs are largely unlinked and
expected to have low LD, so they are excluded from distance-decay estimation.
Within same-chromosome pairs, undefined $r^2$ values (e.g., due to non-variable
loci) are excluded before binning. Distances are binned
into 50 logarithmically spaced intervals, and bins with fewer than 50 SNP pairs
are omitted. Mean $r^2$ $\pm$ SEM is then plotted per retained bin; SEM is
reported as descriptive variability across SNP pairs.

\begin{figure}[H]
\centering
\includegraphics[width=0.93\textwidth]{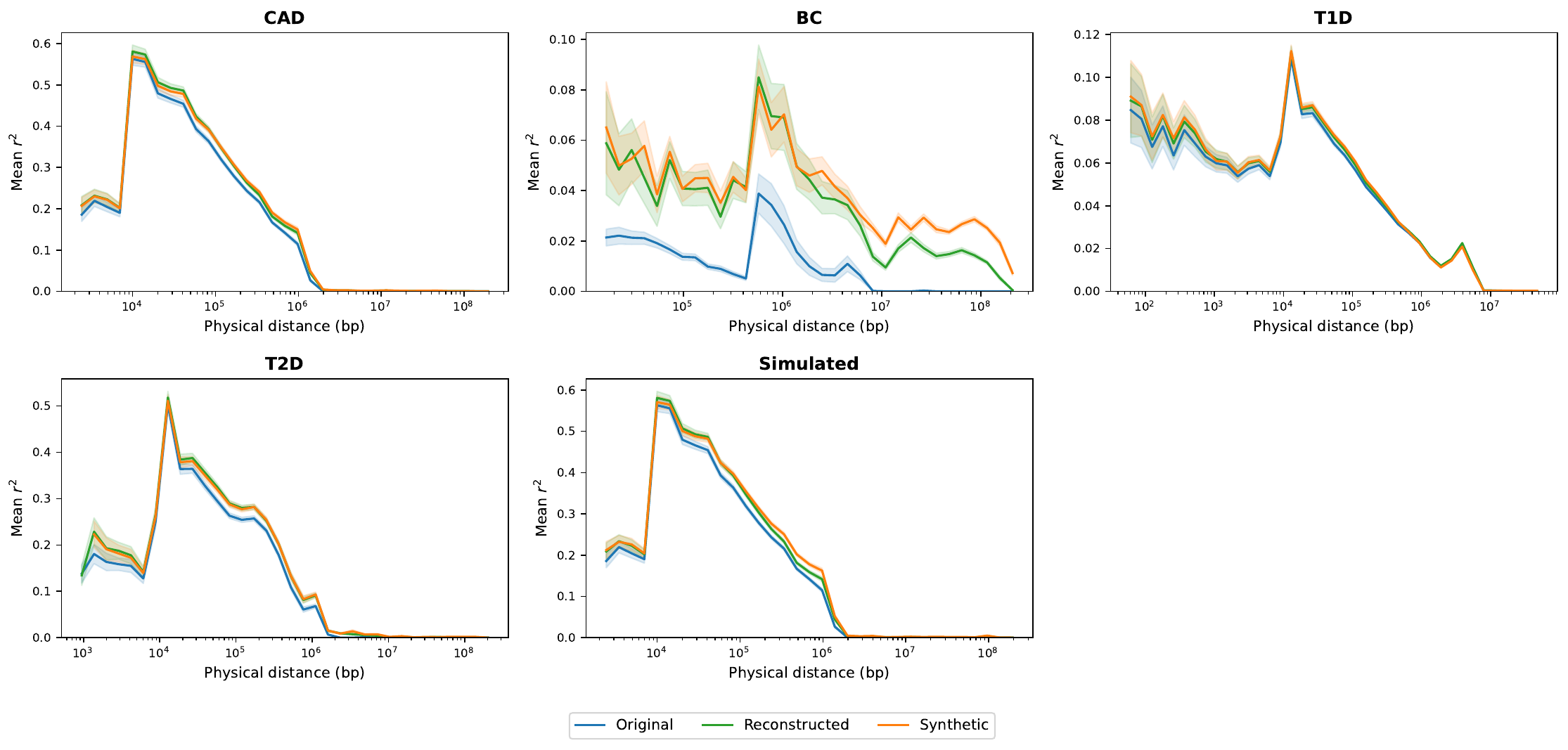}
\caption{LD decay with physical distance for all five traits.  Mean $r^2$
(solid line) $\pm$ SEM (shaded band) is shown for original (blue),
VAE-reconstructed (green), and LDM-generated synthetic (orange) data, computed
over same-chromosome SNP pairs binned into 50 log-spaced distance intervals.\label{fig:supp_ld_decay}}
\end{figure}

Figure~\ref{fig:supp_ld_decay} shows that the LD decay profiles of
reconstructed and synthetic data closely track the original across all traits,
indicating that both the VAE and the LDM preserve the distance-dependent
correlation structure of the real genotype data.
Table~\ref{tab:ld_decay_qc} reports quantitative diagnostics for this analysis.
Spearman correlations between $\log_{10}(\text{distance})$ and $r^2$ are
consistently negative (reflecting the expected decay) and closely matched
between original and generated data for all traits.

\begin{table}[H]
\caption{LD decay diagnostics.  Same-chrom.\ pairs: fraction of all SNP pairs
on the same chromosome (used for decay estimation).  Missing $r^2$: fraction of
same-chromosome pairs with undefined $r^2$ (monomorphic loci).  Spearman
$\rho$: rank correlation between $\log_{10}(\text{distance})$ and $r^2$ over
all valid same-chromosome pairs ($p < 10^{-10}$ for
all).\label{tab:ld_decay_qc}}
{\small
\begin{tabular*}{\textwidth}{@{\extracolsep\fill}llccc@{\extracolsep\fill}}
\toprule
Trait & Dataset & Same-chrom.\ pairs (\%) & Missing $r^2$ (\%) & Spearman $\rho$ \\
\midrule
\multirow{3}{*}{CAD}
 & Original      & 6.7 & 0.0  & $-$0.602 \\
 & Reconstructed & 6.7 & 0.9  & $-$0.638 \\
 & Synthetic     & 6.7 & 0.5  & $-$0.642 \\
\midrule
\multirow{3}{*}{BC}
 & Original      & 6.1 & 0.0  & $-$0.395 \\
 & Reconstructed & 6.1 & 17.2 & $-$0.140 \\
 & Synthetic     & 6.1 & 0.8  & $-$0.096 \\
\midrule
\multirow{3}{*}{T1D}
 & Original      & 93.8 & 0.0 & $-$0.260 \\
 & Reconstructed & 93.8 & 0.1 & $-$0.254 \\
 & Synthetic     & 93.8 & 0.1 & $-$0.260 \\
\midrule
\multirow{3}{*}{T2D}
 & Original      & 6.7 & 0.0 & $-$0.524 \\
 & Reconstructed & 6.7 & 4.8 & $-$0.557 \\
 & Synthetic     & 6.7 & 9.0 & $-$0.566 \\
\midrule
\multirow{3}{*}{Simulated}
 & Original      & 6.7 & 0.0 & $-$0.602 \\
 & Reconstructed & 6.7 & 0.7 & $-$0.631 \\
 & Synthetic     & 6.7 & 0.4 & $-$0.592 \\
\botrule
\end{tabular*}
}
\end{table}

\refstepcounter{suppsec}\label{sec:supp_beta}
\section*{\thesuppsec.\quad Effect Size Preservation: Beta Correlation Analysis}

\begin{figure}[H]
\centering
\includegraphics[width=0.7\textwidth]{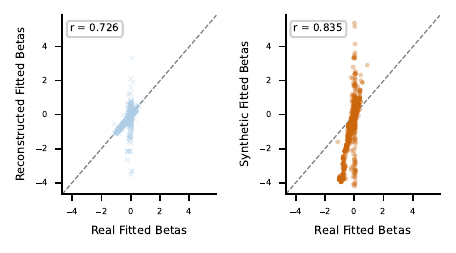}
\caption{Effect size preservation (simulated trait). PRS univariate betas from
reconstructed (left, blue) and synthetic (right, orange) data vs.\ real-data
betas. Synthetic data show higher agreement with real betas (Pearson
$r = 0.835$) than reconstructed data ($r = 0.726$), indicating that
phenotype-conditioned generation preserves marginal associations more
faithfully than unconditional VAE reconstruction.\label{fig:supp_beta_sim}}
\end{figure}

Figure~\ref{fig:supp_beta_sim} compares PRS-univariate effect estimates from
reconstructed and synthetic data against the real-data estimates for the
simulated trait with known causal effects.

\refstepcounter{suppsec}\label{sec:supp_augmentation}
\section*{\thesuppsec.\quad Augmentation Proportion Sensitivity Analysis}

The Syn Augmented condition in the main paper uses the full class-balanced
synthetic dataset, where the number of synthetic samples per class equals the number of real controls. To investigate whether downstream
performance depends on the augmentation proportion---and whether intelligent
sample selection can outperform na\"ive subsampling---we sweep the fraction of
the augmented dataset from 5\% to 100\% under three subsampling strategies:

\begin{enumerate}
\item \textbf{Random subsampling.}  Uniformly samples a given fraction of the
  balanced synthetic dataset.  Repeated with three random seeds (42, 123, 456);
  the mean and standard deviation across seeds are reported.
\item \textbf{Diversity-aware selection.}  Prioritises the most distinct
  synthetic samples within each class.  For each class, every synthetic sample's
  Hamming distance to its nearest same-class neighbour is computed; samples are
  ranked by descending distance and the top-$N$ most diverse are selected first.
\item \textbf{PCA-stratified selection.}  Matches the population structure of
  the real training data.  $K$-means clustering ($k{=}10$) is fitted on PCA
  projections of the real training set; synthetic samples are then drawn from
  each cluster proportionally to the real cluster distribution.
\end{enumerate}

Two baselines are shown for reference: (i)~the non-augmented imbalanced
synthetic dataset (preserving the original class ratio) and (ii)~the
non-augmented balanced synthetic dataset (downsampled majority class).  All
models are evaluated on the held-out real test set using XGBoost with the same
hyperparameter grid as the main experiments (Supplementary Section~\ref{sec:supp_downstream_models}).

\begin{figure}[H]
\centering
\includegraphics[width=\textwidth]{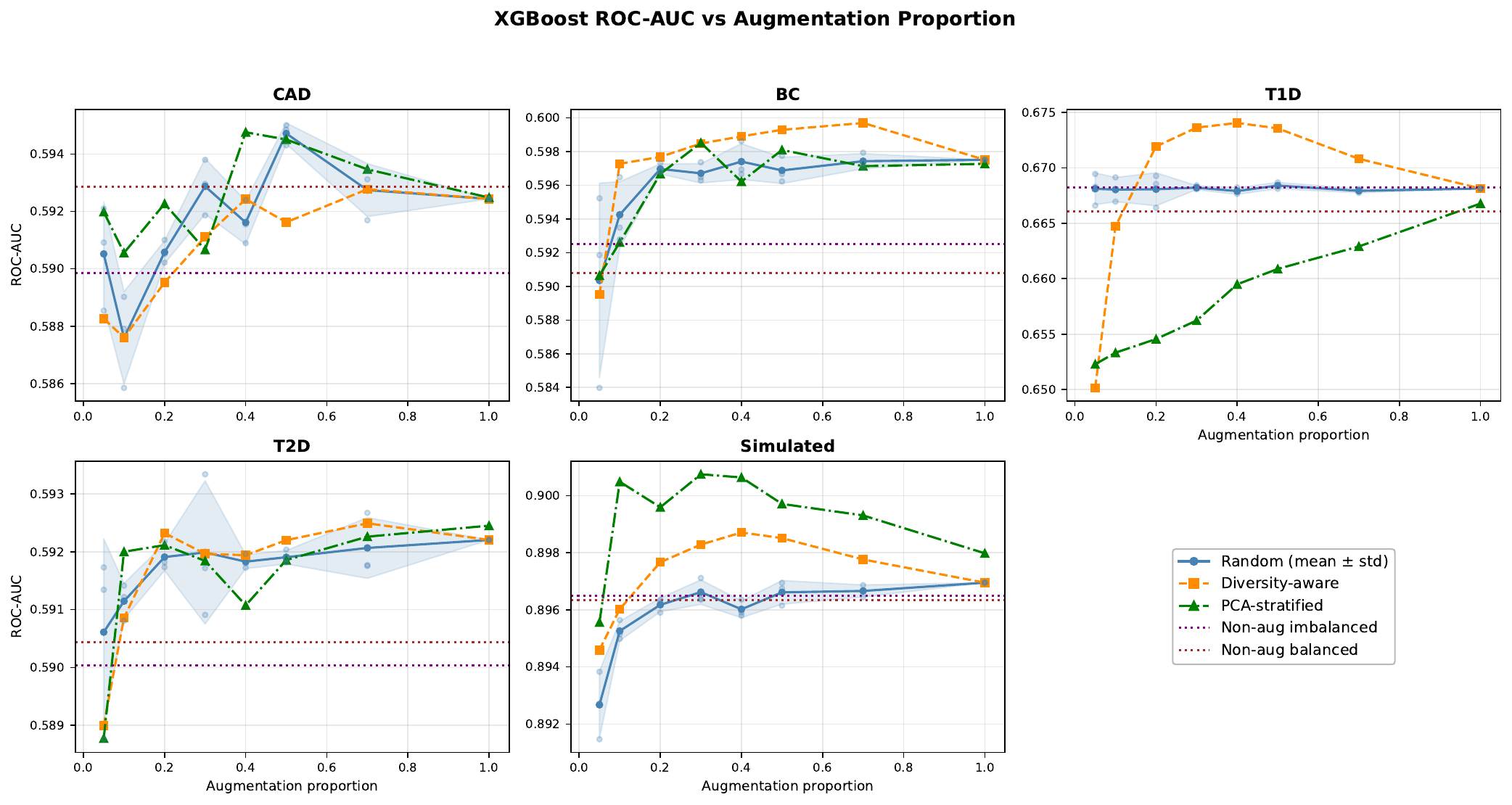}
\caption{XGBoost ROC-AUC on the held-out real test set as a function of
augmentation proportion (fraction of the class-balanced synthetic dataset
used for training) across five traits.  Three subsampling strategies are
compared: random (blue, mean $\pm$ std over 3~seeds), diversity-aware
(orange), and PCA-stratified (green).  Dotted horizontal lines indicate
non-augmented baselines.\label{fig:augmentation_sweep}}
\end{figure}

Figure~\ref{fig:augmentation_sweep} shows the results across all five traits.
Downstream ROC-AUC is largely insensitive to the augmentation proportion for
the four real traits: even at 5\% of the augmented dataset (the smallest
proportion tested), performance is comparable to using the full augmented
dataset.  The three selection strategies yield similar results,
suggesting that the quality of individual synthetic samples matters less
than the overall distributional properties preserved by the generative model.
For the simulated trait, which has a stronger per-SNP signal, a slight upward
trend is observed at higher augmentation proportions,
consistent with the benefit of larger training sets when the signal-to-noise
ratio is favourable.

